\documentclass{article}

 \usepackage[eandd, preprint]{neurips_2026}

\usepackage[utf8]{inputenc} 
\usepackage[T1]{fontenc}    
\usepackage{hyperref}       
\usepackage{url}            
\usepackage{booktabs}       
\usepackage{amsfonts}       
\usepackage{nicefrac}       
\usepackage{microtype}      
\usepackage{xcolor}         
\usepackage{algcompatible}
\usepackage{algorithm}
\usepackage{amsmath}
\usepackage{graphicx}
\usepackage[table]{xcolor}
\definecolor{lightgray}{gray}{0.92}
\usepackage{wrapfig}
\usepackage{tabularx}
\usepackage{multirow}
\usepackage{makecell}
\usepackage{multicol}
\usepackage[skins,breakable, most]{tcolorbox}
\usepackage{hyperref}
\usepackage{fontawesome}
\usepackage{ragged2e}

\definecolor{pink-color}{RGB}{237,46,104} 
\newcommand{\paramnorm}[1]{
\textcolor{pink-color}{\small{\texttt{\detokenize{#1}}}}
}
\newtcolorbox[list inside=prompt,auto counter,number within=section]{prompt}[1][]{
    colbacktitle=black!80,
    colframe=black!80,
    coltitle=white,
    fontupper=\footnotesize,
    boxsep=5pt,
    left=0pt,
    right=0pt,
    top=0pt,
    bottom=0pt,
    boxrule=1pt,
    enhanced, 
    breakable,
    skin first=enhanced,
    skin middle=enhanced,
    skin last=enhanced,
    #1,
}

\usepackage[textsize=tiny]{todonotes}

\title{\raisebox{-9pt}{{\includegraphics[height=28pt]{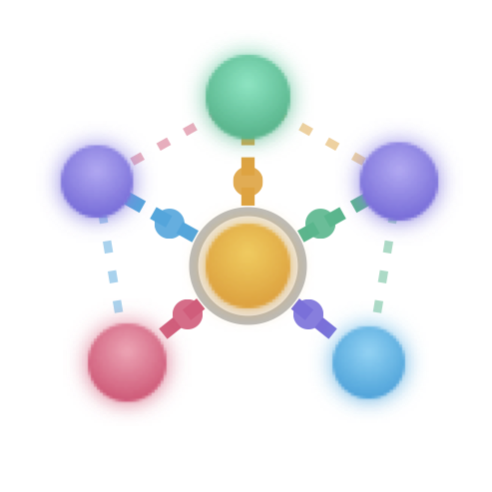}}} CREATE: Testing LLMs for\\ Associative Creativity}

\author{%
  Manya Wadhwa\thanks{Corresponding Author: mw4141@nyu.edu} \\
 New York University
  \And
  Tiasa Singha Roy \\
  New York University \\
  \AND
  Harvey Lederman \\
  The University of Texas at Austin \\
  \And
  Junyi Jessy Li \\
The University of Texas at Austin \\
  \And
Greg Durrett \\
  New York University \\
}

\begin{document}

\maketitle

{
\hypersetup{urlcolor=black, pdfborder={0 0 0}}
\vspace{-2em}
\begin{center}
\centering
\href{https://manyawadhwa.github.io/projects/create/}{%
  \raisebox{-2pt}{\includegraphics[height=13pt]{logos/create-logo3.png}}%
  \; \textbf{Project Page}%
}
\quad\quad
\href{https://github.com/ManyaWadhwa/CREATE}{%
  \raisebox{-1pt}{\faGithub}\; \textbf{Repository}%
}
\quad\quad
\href{https://huggingface.co/datasets/wadhma/CREATE}{%
  \raisebox{-2pt}{\includegraphics[height=11pt]{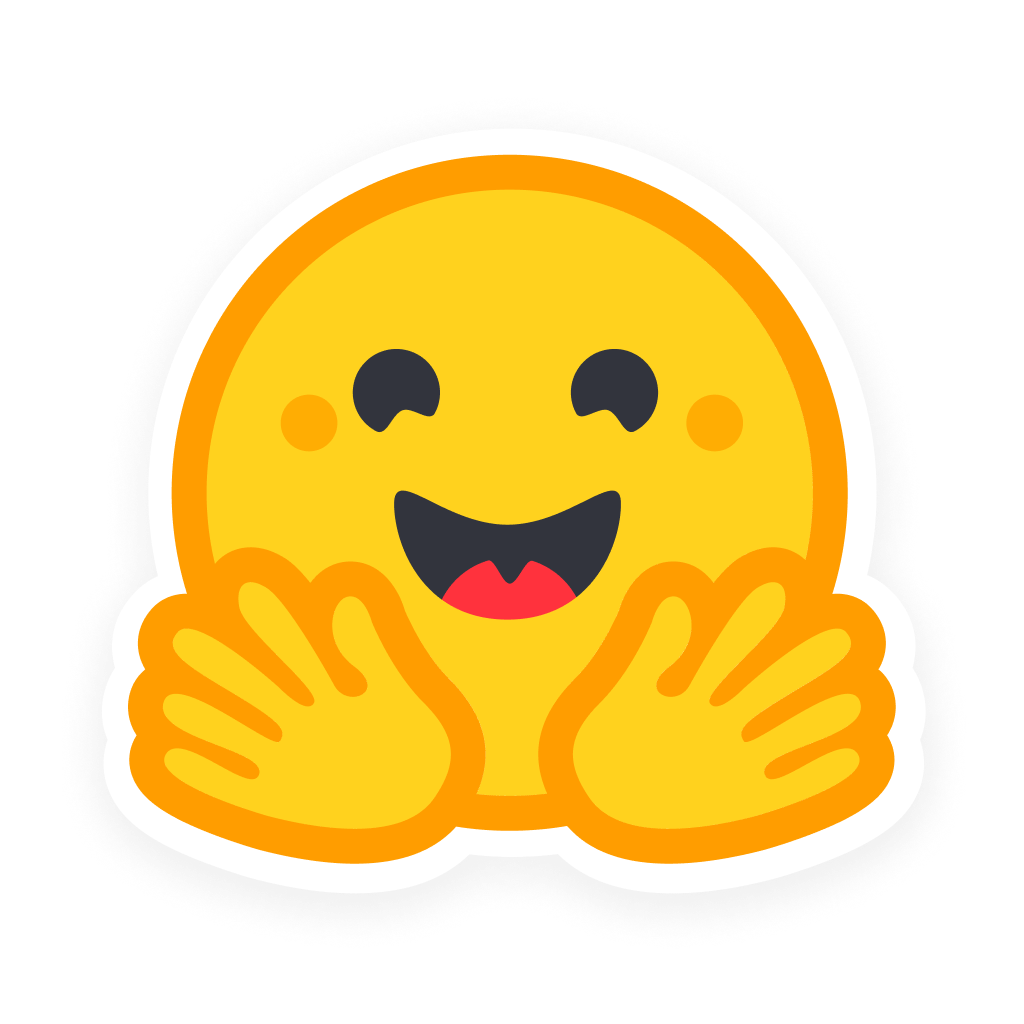}}\; \textbf{Dataset}%
}
\end{center}
}

\vspace{2em}

\begin{abstract}
A key component of creativity is associative reasoning: the ability to draw novel yet meaningful connections between concepts. We introduce CREATE, a benchmark designed to evaluate models' capacity for creative associative reasoning. CREATE requires models to generate sets of paths connecting concepts in a model's parametric knowledge. 
Paths should have high \textit{specificity} (distinctiveness and closeness of the concept connection) and high \textit{diversity} (dissimilarity from other paths), and models are scored more highly if they produce a larger set of strong, diverse paths. This task shares demands of real creativity tasks like hypothesis generation, including an extremely large search space, but enables collection of a sizable benchmark with objective answer grading. Evaluation of frontier models shows that the strongest models achieve higher creative utility than others, but high multiplicity of answers and complexity of the search makes benchmark saturation difficult to achieve. Furthermore, our results illustrate that thinking models are not always more effective on our task, even with high token budgets. Recent approaches for ``creative prompting'' give some but limited additional improvement. CREATE provides a sandbox for developing new methods to improve models' capacity for associative creativity.
\end{abstract}

\section{Introduction}

Creativity, as one of the three pillars in Sternberg's Triarchic theory of intelligence \cite{sternberg1985beyond} and the highest cognitive skill in Bloom's Taxonomy \cite{bloom1964taxonomy}, is central to scientific discovery, writing, and creative problem solving. A flurry of recent work aims to develop AI agents for these tasks; for example, for hypothesis generation \cite{liu2025hypobench}, research idea generation \cite{guo2025ideabench}, and the entire scientific process \cite{Lu2024TheAS,majumder2024discoverybench,Mitchener2025KosmosAA,agarwal2025autodiscovery,si2026executiongroundedautomatedairesearch}. But how do we know if our models are creative enough? Real-world, complex queries are challenging and subjective to evaluate \cite{chakrabarty2024art,wadhwa2025evalagent}, while symbolic benchmarks using abstract tasks \cite{nagarajan2025roll, schapiro2025combinatorial} may not reflect how the models are used in reality.

This work presents a new benchmark for \textbf{associative creativity}, striking a balance between real-world applicability and verifiability. 
We take motivation from \textit{combinatorial creativity} \cite{10.5555/102753}, which requires the formation of new thoughts by linking, recombining, or drawing analogies across familiar concepts. This process closely aligns with what we term associative creativity and plays a foundational role in creative reasoning.
Prior research, from Koestler’s bisociation \cite{koestler1964act} to Gentner’s structure-mapping theory \cite{gentner1983structure}, demonstrates that many creative insights originate in the \textbf{associative leaps} characteristic of combinatorial creativity. 
Our work focuses on associations between \emph{concepts}: being able to surface nonobvious but striking connections between known concepts is crucial for new discoveries.

\begin{figure}
    \centering
    \includegraphics[scale=0.22,trim=1cm 15cm 5cm 5cm]{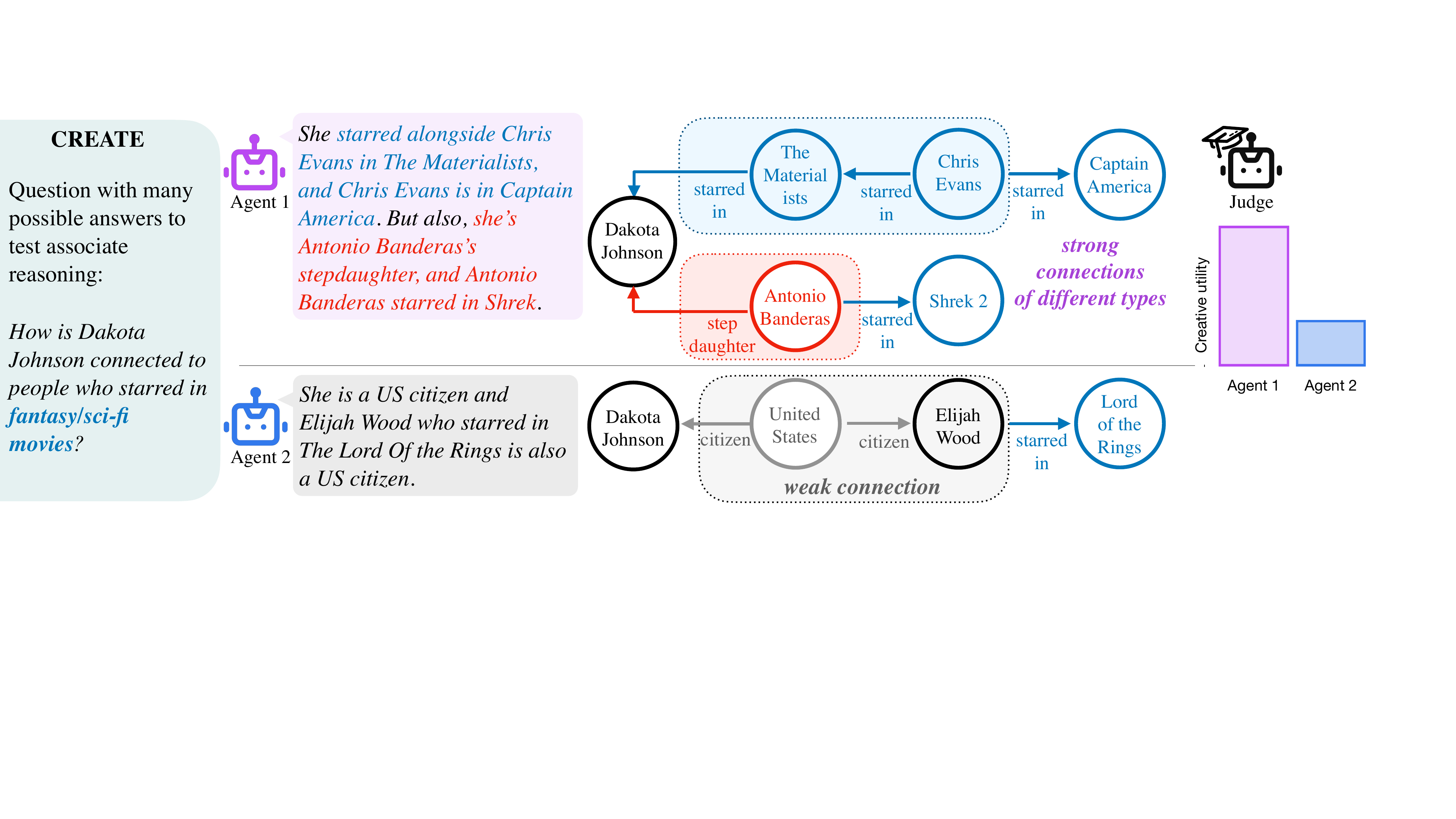}
    \caption{Motivating example of brainstorming paths in knowledge graphs. In CREATE, only the question is given; reasoning over the graph is implicit in the model's parameters and thinking trace, similar to drawing connections for scientific research. Agents are rewarded in our \emph{creative utility} metric for finding more strong, distinct paths.}
    \vspace{-1em}
    \label{fig:intro}
\end{figure}

We present CREATE, a benchmark designed to evaluate LLMs’ ability to form open-ended associations among real-world entities.
Figure~\ref{fig:intro} shows an example of a query designed to test this kind of reasoning. CREATE is \textit{practical}, as it relies on world knowledge about concepts and relations rather than synthetic scenarios; it also sits at the \textit{right level of complexity}: it is simple enough to evaluate yet still requires multi-hop reasoning about real concepts.

The queries in CREATE are constructed to span different conceptual domains: people in movies, people in positions of power, genes associated with certain diseases etc. By formulating questions in a sufficiently broad manner, we can test whether LLMs can enumerate the creative, ``trivia-like'' connections that a human domain expert might be aware of as salient and interesting. Figure \ref{fig:intro}'s possible answers show diverse and interesting possible connections. Our task involves reasoning about a more open-ended answer space compared to past work on multi-hop reasoning like HotpotQA \cite{yang2018hotpotqa} and similar datasets \cite{trivedi2022musique,he2024mintqa,ding2024knowledge}. However, CREATE has both objective and hard-to-find answers, unlike more ``brainstorming''-oriented tasks \cite{zhang2025noveltybench}. The ability to enumerate large numbers of distinct, strong connections directly ties to the ability to do strong brainstorming in other ideation settings.

Our central goal is for LLMs to generate connections that (a) reflect close but non-typical associations between concepts and (b) are diverse rather than simple variations on a single theme. These criteria align with standard desiderata for creative products: high quality coupled with meaningful novelty. Following prior work \cite{zhang2025noveltybench}, we capture both aspects with a unified \textit{creative utility} metric that integrates quality and diversity. 

Our results show that frontier models achieve higher creativity utility scores compared to weaker models or open-source models. However, when we zoom in on particular cases, we find that no single model adequately covers the space of strong answer paths. 
Our analysis also shows that spending more reasoning tokens does not necessarily lead to higher scores on our benchmark, raising questions about how current AI systems search conceptual spaces and what is required for them to genuinely support or participate in creative processes. 

\section{Background} \label{sec:background}
\paragraph{Tests for human creativity and diversity} Human creativity is often assessed using tasks such as the Alternative Uses Task \cite{guilford1978alternate}, Divergent Association Task \cite{olson2021naming}, and Remote Associates Test \cite{mednick1968remote}. While these are demanding for humans, taxing working memory and requiring the coordination of multiple concepts under cognitive constraints, they are comparatively easy for LLMs \cite{wenger2025we}, which can exploit large context windows, retrieval tools, and parallel generation. In addition, standard test instances are often seen during pre-training, unlike for human participants. We show the saturation of these benchmarks more in Appendix~\ref{sec:rat_aut_etc}. By contrast, our benchmark requires forming novel connections among real-world entities (\textit{Sebastian Vettel, DSG3 }) rather than familiar everyday objects (e.g., \textit{book}, \textit{bottle}, \textit{brick}) that are commonly used in the Alternative Uses Task.

\paragraph{Tests of model creativity} Prior work proposes benchmarks \cite{ruan2024liveideabench, lu2025benchmarking, padmakumar2025beyond, fein2025litbench, guo2025ideabench, liu2025hypobench}  focusing on specific creative abilities like creative writing, scientific ideation and code generation. However, evaluation in these specialized domains is challenging where the use of LLM-as-a-judge for studies at scale can conflate quality, diversity and usefulness \cite{si2025ideation}. It is unclear how to scale human evaluation: even AI-created papers undergoing real review processes are later found to have major flaws \cite{liu2025promisespitfalls}. In contrast,  Nagarajan et al.~\cite{nagarajan2025roll} and Schapiro et al.~\cite{schapiro2025combinatorial} introduce symbolic, highly controlled testbeds for probing combinatorial creativity and training models on abstract tasks. However, these settings are far removed from the complex, open-ended ways LLMs are used in real applications. Our work achieves the best of both worlds:
we propose a realistic, knowledge-grounded task involving real-world entities and concepts. By leveraging knowledge graphs and familiar domains, we expect LLM-as-a-judge evaluations to be comparatively reliable.

Models are also criticized for returning homogeneous outputs \cite{zhang2024forcing,jiang2025artificial}. We do not directly engage with ``distributional pluralism'' \cite{sorensen2024roadmappluralisticalignment,lake-etal-2025-distributional,west2505base} in this work; however; it indicates that LLMs may face a performance ceiling on our benchmark, even if ensembled together, 
and demonstrates why current LLMs may not be creative enough in practice \cite{anderson2024homogenization, padmakumar2309does}. 

\section{Conceptual Framework} \label{sec:framework}

\subsection{Defining and measuring associative creativity} \label{subsec:creativity-definition}

Define $\mathcal{U}(\mathbf{x})$ as a set of possible outputs for some task $\mathbf{x}$. For instance, if $\mathbf{x}$ is a mathematical theorem, $u \in \mathcal{U}(\mathbf{x})$ is a proof, if $\mathbf{x}$ is a writing prompt, then $u$ is a story, etc.

We assume an LLM can produce a set $U \subset \mathcal{U}(\mathbf{x})$ through some sampling procedure, which we denote via $U \sim \pi_{\mathrm{LLM}}(U \mid p(\mathbf{x}))$ for a prompt $p$ conditioned on the task. A key feature of our tasks is that $|U|$ is likely to be of moderate size at most (10-100) with standard use of LLMs; an LLM must spend significant compute to identify high-quality items $u$, so enumerating large numbers of possible $u$ is challenging.

We evaluate on two dimensions. First, a \textbf{quality measure} $f: u \in \mathcal{U} \rightarrow \mathbb{R}$ assigns a score to each item $u$. Second, a \textbf{distance measure} $d: \mathcal{U} \times \mathcal{U} \rightarrow \mathbb{R}_{\geq 0}$ defines distances between elements of $\mathcal{U}$.\footnote{We note that these criteria are explored in the theory of submodular functions \cite{lin-bilmes-2011-class} and in techniques like determinantal point processes \cite{Kulesza2012DeterminantalPP}}

We can combine these objectives into a single metric following NoveltyBench \cite{zhang2025noveltybench}. We define the \textbf{creative utility} of a set $U$ as 
\vspace{-1em}

\begin{equation} \label{eq:utility}
s(U) = \max_{\tau} \sum_{i=1}^{|U|} \gamma^{i-1} f\!\left(u_{\tau(i)}\right) \min_{j<i} d\!\left(u_{\tau(i)}, u_{\tau(j)}\right),
\end{equation}

where $\tau$ denotes an ordering over $U$. Items are scored via their incremental marginal utility under this ordering, which is a function of how many items have already been selected ($\gamma$), their distance to previously selected items ($d$), and their quality $f$. The first item is assumed to have a distance of 1 from the empty set. 

$\gamma$ defines a user ``patience.'' Values close to 1 favor larger sets, which we believe reflects how LLMs would be used for challenging creative tasks in practice: a user could scan through a moderately-sized list of ideas.\footnote{This objective differs from that of NoveltyBench in two key ways. First, their notion of item redundancy is discrete 0-1 rather than continuous, which is not appropriate for our setting. Second, we eliminate their normalization term $\frac{1-\gamma}{1-\gamma^{|U|}}$; $\gamma$ already discounts additional items and this terms tends to strongly penalize returning larger sets.}

We assume that LLMs return unordered sets $U$ (e.g., through repeated sampling), but $s(U)$ requires an ordering to be computed. We compute a greedily optimal $\tau_i$ by finding the element $u_k$ that maximizes
\begin{equation}
f(u_k)\min_{j < i} d(u_k,u_{\tau_j})
\end{equation}
\vspace{-1.5em}



\paragraph{Associative creativity on graphs} This work focuses on associative creativity, which instantiates $\mathcal{U}$ as paths in graphs. 
At a high level, many creative insights come from linking concepts through structured relationships. For example, a drug and a disease may be connected through shared biological pathways or intermediate processes, rather than by a single direct link. This abstraction provides a simple way to study the associative part of combinatorial creativity. Concepts or entities in our graphs are nodes and relationships between them are edges. From this view, generating a creative association means finding a high-quality (strong) connection, and ideally one that is non-obvious. 

One distinguishing feature of this setting is that there are potentially large numbers of valid paths. However, quality is right-tailed: that is, there are elements of $\mathcal{U}$ that are substantially more valuable than others, but which are hard to find. We can think of these as works of great literature, or great creative ideas, amidst a large number of more conventional (but still well-executed) concepts. This differs from NoveltyBench, where many responses to a prompt (\emph{What's the best car to get in 2023?}) may be equivalent in quality.

\begin{wrapfigure}{r}{0.6\textwidth}
\vspace{-3em}
\centering
\includegraphics[scale=0.22, trim= 0 19cm 18cm 0]{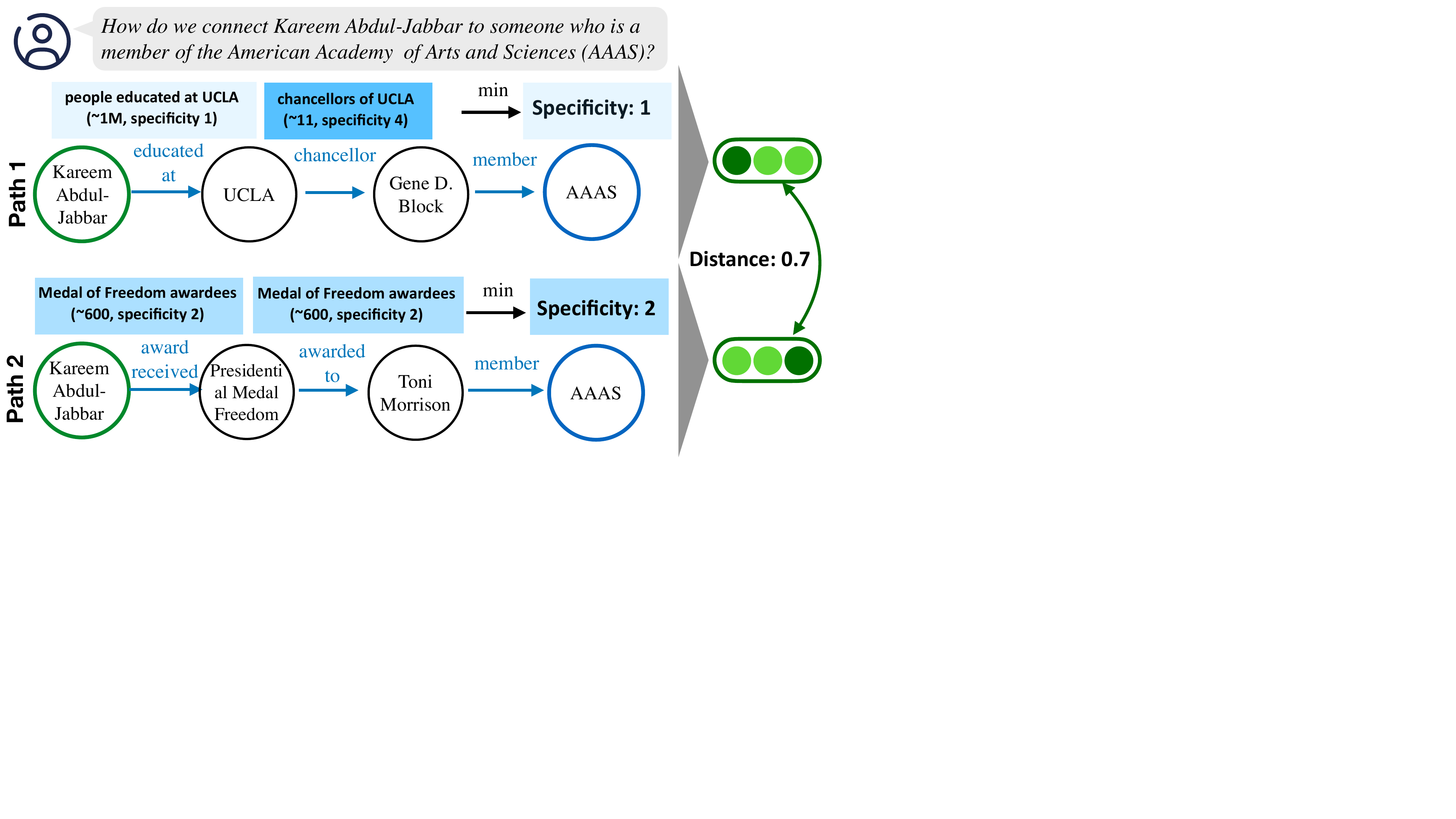}
\caption{Scoring generated paths. For specificity, we  prompt an LLM as a judge to give an estimated class size, which is mapped to a number between 1-5. Triples belong to a large class size get a low score. The specificity of a path is the specificity of its weakest triple. Paths are embedded and compared to compute distance}
\label{fig:strength_example}
\vspace{-1em}
\end{wrapfigure}

\subsection{CREATE Formalization} \label{subsec:formal}
We now concretize our framework in the setting of reasoning about graph connections.
Let $G = (E,R)$ be a knowledge graph with entities $E$ and relations $R$. Paths $u$ are represented by sequences of triples $u = \big[(e_1, r_1, e_2), \ldots, (e_n, r_n, e_{n+1})\big]$
where each triple $(e_i, r_i, e_{i+1})$ consists of two entities $e_i, e_{i+1} \in E$ connected by a relation $r_i \in R$. We say that a path is \textbf{structurally valid} if consecutive triples share entities, i.e., the second entity of $(e_i, r_i, e_{i+1})$ equals the first entity of $(e_{i+1}, r_{i+1}, e_{i+2})$ (Note that LLMs may produce structurally invalid paths in their generations). We say that a path is 
\textbf{factual} if the triples represent true relations.

The universe $\mathcal{U}(\mathbf{x})$ consists of all structurally valid paths in $G$ connecting entities $e_i,e_k \in E$ and obeying the constraints of the prompt $\mathbf{x}$. We evaluate factuality separately.

Our prompts $\mathbf{x}$ go beyond simple queries and instead ask about \textit{sets}, as shown in Figure~\ref{fig:intro}. 
We only consider structurally valid paths that 
satisfy the prompt given by $\mathbf{x}$; in Figure~\ref{fig:intro}, whether a path successfully connects Dakota Johnson with a fantasy or sci-fi movie.

\paragraph{Quality of Paths} 
We say that paths are high-quality if they are factual and if they are specific.

Specificity measures how many entities could plausibly participate in a given connection. Informally, we would think of \textit{Dakota Johnson} as more strongly connected to \textit{Antonio Banderas} than she is to other actors (Fig.~\ref{fig:intro}); she has fewer step-fathers than co-stars. Being connected by being citizens of the same country is a weak or a less specific connection, unless it were a particularly small country.

This intuition suggests that what governs the specificity of a relationship is the number of entities that typically participate in it on a given side. More precisely, since paths are composed of triples, we define the quality of a \emph{path} as the specificity of its weakest triple.

For an individual triple $(e_i, r_i, e_{i+1})$, we define two predicate-induced classes:
\begin{align}
\mathcal{C}_A(e_i, r_i, e_{i+1}) 
&= \{\, x \mid (x, r_i, e_{i+1}) \text{ is true} \,\},
\mathcal{C}_B(e_i, r_i, e_{i+1}) 
&= \{\, y \mid (e_i, r_i, y) \text{ is true} \,\}.
\vspace{-0.5em}
\end{align}

The specificity of a triple is then defined as a function of the larger of these two classes:
\begin{equation}
\label{eq:strength}
\small
\sigma(e_i, r_i, e_{i+1}) 
= g\!\left( 
\max\big(
|\mathcal{C}_A(e_i, r_i, e_{i+1})|,\;
|\mathcal{C}_B(e_i, r_i, e_{i+1})|
\big) 
\right).
\end{equation}

where $g(\cdot)$ is a monotonically decreasing function, so that triples participating in large, non-selective classes receive lower specificity scores. For example, in Figure \ref{fig:strength_example}, specificity scores for the triples in the first path are [1,4] and thus the score for the path is 1. 

Factuality ensures that relations generated in $u$ exist in the graph. Let $q(u)$ denote a factuality function for path $u$, where $q(u)=1$ if all triples in $u$ represent true relations.

The quality $f(u)$ of a path $u$ (from  Section ~\ref{sec:framework}) is defined as
\begin{equation}
\label{eq:quality}
f(u)
=
\mathbb {I}[ q(u) =1 ] \min_{(e_i, r_i, e_{i+1}) \in u}
\sigma(e_i, r_i, e_{i+1}).
\end{equation}
Intuitively, a path is only as strong as its weakest relation. This formulation favors paths composed of selective, informative relations and penalizes those with 
highly generic edges.

Both specificity and factuality are implemented with LLM judges described in Section~\ref{sec:benchmark}. These judges are evaluated for quality in Section~\ref{sec:case_study}.

\paragraph{Distance function}
We define the following distance function to be used in the measures in Section~\ref{subsec:creativity-definition}.
Let a path $u$ be represented by its string form $\mathrm{str}(u)$. We define the distance between two paths as a transformation function of the cosine distance between their string representations. 
\vspace{-0.5em}
\begin{equation}
 d(u_i,u_j)
=
g (
1-\cos(\mathrm{str}(u_i), \mathrm{str}(u_j))),
\end{equation}
where, $g(x)$ is a cosine-annealed cosine distance 
\begin{equation}
g(x)=
\begin{cases}
\frac{1}{2}\left(1-\cos\left(\pi (\frac{x}{0.7})^{2}\right)\right) & 0 \le x \le 0.7, \\[6pt]
1 & 0.7 < x \le 1.
\end{cases}
\end{equation}
We use this transformation because we observed that paths with distances $>0.7$ differed very substantially, with incidental overlap due to a few shared tokens or concepts within the domain. Paths with distances $<0.4$ were almost always varying on very similar connections; we did not perceive any marginal value from these. Our function rescales the distance interval to account for these judgments; see plot and examples in Appendix ~\ref{app:distance}.


\section{Constructing the CREATE Benchmark} \label{sec:benchmark}
\textbf{Dataset Curation:} 
We use Wikidata to construct queries in CREATE.
The data generation process is described in Algorithm~\ref{alg:nlq-generation} in Appendix ~\ref{app:benchmark}. We begin by manually selecting a set of relation–category pairs $(r,c)$, each of which defines a class
$C_{r,c} = \{ x \mid (x, r, c) \in G \}$. These pairs are chosen to ensure that the resulting classes are compact and semantically coherent, with members linked by a specific role, condition, or function, for example, one $(r,c)$ pair in our dataset is (\textit{member of sports team, Scuderia Toro Rosso)}. Triples belonging to the corresponding class $C_{r,c}$, sampled from $G$, include \textit{[(Sebastian Vettel, member of sports team, Scuderia Toro Rosso), (Daniel Ricciardo, member of sports team, Scuderia Toro Rosso), (Alex Albon, member of sports team, Scuderia Toro Rosso) , $\ldots$, (Max Verstappen, member of sports team, Scuderia Toro Rosso)]}.


In the next step, we form unordered pairs from entities in $C_{r,c}$ within the class. For each pair, we randomly select one of the two triples and sample informative outgoing edges from $G$, yielding an additional one-hop relation. For example, we create a pair: \textit{[(Sebastian Vettel, member of sports team, Scuderia Toro Rosso), (Max Verstappen, member of sports team, Scuderia Toro Rosso)]}, and expand on the second triple to find the link: \textit{(Max Verstappen, unmarried partner, Kelly Piquet)}. We then create a source path that links the three together in a chain: \textit{[(Sebastian Vettel, member of sports team, Scuderia Toro Rosso), (Max Verstappen, member of sports team, Scuderia Toro Rosso),  (Max Verstappen, unmarried partner, Kelly Piquet)]}. The resulting query is formed using the head of the first triple and the tail entity of the last triple. We use {\small{\texttt{gpt-4o-mini-2024-07-18}}} to re-write the structured triple into a natural language query: \textit{``What is the connection between Sebastian Vettel and someone who has been/is a partner of Kelly Piquet?''} We show additional examples in Table \ref{tab:benchmark}. 

Note, we use this source path only as a way to establish that there exists at least one strong path between the entities. \textbf{We do not evaluate against it as a ground truth reference}; we instead evaluate a model on its ability to generate multiple distinct and strong paths. 

\textbf{Benchmark Statistics:}
We select eleven relations ($r$) that span diverse domains. The starting entity is one of the types: people, genes, or chemical compounds, spanning both encyclopedic knowledge, classic ``trivia'' domains like movies, and scientific knowledge. In total we have 931 natural language queries, with at least one verified connection between the entities. Table~\ref{tab:relation_stats} lists the relations, number of queries per relation and number of unique starting entities. Examples of source paths and generated queries are given in Table~\ref{tab:benchmark}; examples are also shown in Figure~\ref{fig:case_study}. We describe the filtering process in Appendix \ref{app:benchmark}.


\textbf{Quality Metric $f(u)$}: We estimate triple specificity using a single prompt in which a large language model ({\small\texttt{gpt-oss-120b}}) jointly assesses the sizes of both predicate-induced classes and directly identifies the larger of the two. This estimated maximum class size is mapped to a discrete score on a five-point scale. Appendix~\ref{sec:eval_prompts} provides more details. We evaluate factuality using an LLM-as-a-judge ({\small\texttt{gpt-oss-120b}}) on the relation level. The prompt for evaluating factuality is given in Appendix \ref{prompt:factuality_prompt}.

\textbf{Distance Metric $d(u_i, u_j)$}: The pairwise distance between paths, by getting their embeddings using {\small\texttt{all-MiniLM-L6-v2}} and using the transformed cosine distance as the function. 



\vspace{-0.5em}

\section{Experimental Setup}
\label{sec:experimental_setup}

\textbf{Models:} We evaluate both non-thinking and thinking large language models on the proposed task, covering a range of architectures, sizes, and training paradigms.
\textbf{Non-thinking models} include GPT-4.1-mini, GPT-4.1, Qwen2.5-32B-Instruct \cite{qwen2.5}, Qwen3-30B-A3B-Instruct-2507\cite{qwen3technicalreport}, and OLMo-3.1-32B-Instruct \cite{olmo2025olmo3}.
\textbf{Thinking models} include GPT-5.5, GPT-5.4, GPT-5, GPT-5-mini, Gemini-3.1-Pro, Gemini-3-Pro, Claude-Opus-4.7, Claude-Sonnet-4.6, Claude-Haiku-4.5, Qwen3-32B \cite{qwen3technicalreport}, OLMo-3.1-32B-Think \cite{olmo2025olmo3} and gpt-oss-120b \cite{openai2025gptoss120bgptoss20bmodel}. Appendix \ref{app:model_details} gives model and inference details.

\textbf{Base prompt:} All models are evaluated using a shared base prompt (Appendix~\ref{prompt:base_prompt}) (referred to as `original' in the results) designed to elicit diverse, high-quality relational connections.   We prompt the model to generate multiple, possible candidates, so we end up with a set of connections as the output. The model is instructed to output these multiple paths in the form a JSON enclosed by $<$answer$>$ tags. We then parse out the text into a structured JSON to get all model generated paths. To ensure comparability across models, we use a temperature of 0.7 and a maximum generation length of 4096 tokens for non-thinking models. For thinking models we found this max length was too small, so we vary the token budget/available reasoning budget. For the frontier models that are API-based, we test with the default reasoning budget (unless stated otherwise), with the exception of GPT-5-mini, where we vary the reasoning effort parameter across the supported settings (\textit{low}, \textit{medium}, and \textit{high}). For the open-source models, we explicitly vary the reasoning budget (16k and 32k). Finally, we filter all paths to be structurally valid paths (Section \ref{sec:framework}) before computing results.

\textbf{Prompt Variations:} One potential approach to mitigating lack of creativity is the use of alternative prompting strategies. Prompt variations are commonly employed to steer model behavior and elicit alternative responses, for example, by paraphrasing the prompt or explicitly requesting a different answer when the initial output is unsatisfactory \cite{zhang2025noveltybench}. In our experiments, we examine whether such prompt-based interventions are effective in eliciting more diverse and higher-quality relational paths from large language models.

\textbf{(1) Be creative.} We augment the base prompt with the instruction: \textit{``Be creative in the type of relationships explored and generated.''} to encourage broader exploration of relation types.

\textbf{(2) Verbalized Sampling.} We adopt verbalized sampling \cite{zhang2025verbalized} to reduce mode collapse by having the model explicitly express a probability distribution over its outputs. Detailed prompt is given in Appendix~\ref{prompt:verbalized_sampling}.

\textbf{(3) In-context regeneration/iterative}: After the initial generation, we query the model once again and explicitly ask the model to provide a different answer while keeping all previous answers in the conversation context. This allows the model to see its previous responses and deliberately generate something new. The prompt for this is given in Appendix ~\ref{prompt:iterative}.

\textbf{(4) Resampling}: We obtain multiple independent generations from the model using the base prompt (Appendix \ref{prompt:base_prompt}) with a temperature set to 0.7. Each generation is independent and the model has no knowledge of previous generations.

\section{Metric Validation} \label{sec:case_study}


First, we ensure that our evaluation procedure appropriately measure specificity, factuality, and creativity of our paths. We conduct a series of human validation experiments along these axes.

\textbf{Specificity:} First, we validate that the LLM judge's specificity scores align with human judgments. The LLM judge produces a class-size estimate mapped to a 1–5 score. Two authors independently annotated 75 paths, assigning scores on the same 1–5 scale while keeping the underlying class-size mapping in mind. The annotators agree substantially, achieving a Krippendorff's alpha \cite{krippalpha} of 0.68. The Pearson correlation between the average of the two annotator ratings and the {\small\texttt{gpt-oss-120b}} ratings is 0.67, indicating agreement between the judge and the annotators as well. Some of the main disagreements are because of differences in knowledge domains between annotators and their familiarity with the entities.

\begin{figure*}[t!]
    \centering
    \includegraphics[width=\linewidth]{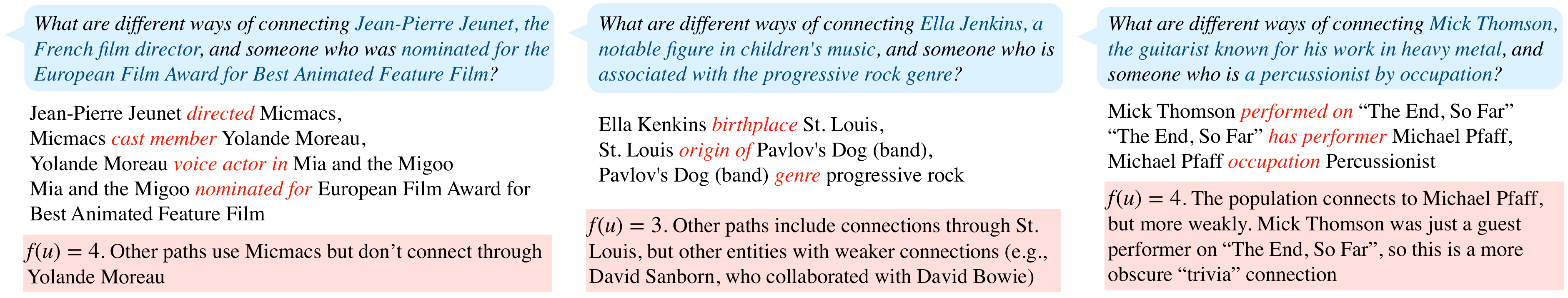}
    \vspace{-1em}
    \caption{Examples of model-generated paths along with quality scores. Path scores are intuitively sensible, and these paths are unique to a single model despite seeming interesting. Table \ref{tab:case_study_app} shows the corresponding closest population paths along with the distance numbers.}
    \label{fig:case_study}    
\end{figure*}


\textbf{Factuality:}
To validate the judge for factuality, one of the authors annotated a set of 346 model-generated triples. The annotator used web search to try to substantiate each relation and produce high-quality labels. On this human-labeled set, the factuality judge achieves an overall balanced accuracy of 85.9\%. For incorrect relations, the evaluator achieves high recall (0.94) but lower precision (0.52), while for correct relations it achieves high precision (0.98) and recall (0.77). The most common sources of error are entity/relation misinterpretation and knowledge of long-tail entities. 

\textbf{Robustness to judge model:} Recall that both factuality and specificity are evaluated using {\small\texttt{gpt-oss-120b}}. Tables \ref{tab:factuality_diff_models} and \ref{tab:specificity_diff_models} show that our choice of LLM-as-a-judge is comparable to stronger models, and Table \ref{tab:stability} confirms that our main findings are robust to this choice. \textbf{Taken together, these results demonstrate that LLM judges for specificity and factuality (1) correlate substantially with human ratings; (2) are robust enough for model rankings to remain unchanged when the judge model is changed.}


\textbf{Creative Utility:} Having established that specificity and factuality are appropriately measured, we now evaluate our metric end-to-end. Two authors independently scored model-generated outputs for 105 CREATE queries on the basis of \emph{overall creativity}. These judgments were anchored to a 1-5 scale with rubric guidance, but ultimately rest on a subjective notion of creativity similar to the analysis in Figure \ref{fig:case_study}: does the connection seem salient and interesting as a piece of trivia? This evaluation is inherently less quantitatively grounded than the other metrics in this paper, and it relies on the annotators' domain knowledge, which is not uniform across the represented queries.

The annotators achieved a Krippendorff's alpha of 0.62, indicating moderate agreement in their relative rankings. The Pearson correlation between each annotator and the creative utility metric was 0.51 and 0.48 respectively, with the correlation between annotator averages and the metric was 0.55. 

\textbf{Crucially, end-to-end scoring with a judge is not a viable basis for our benchmark.} We show in Appendix \ref{subsec:stability} that our proposed metric produces stable rankings across evaluators, compared to directly prompting an LLM with a holistic scoring rubric, supporting the value of decomposition over end-to-end LLM judgment for creativity.

Taken together, these results show that (1) our evaluations of specificity and factuality appropriately measure these quantities; (2) the overall creativity utility metric correlates well with human judgments; (3) the creative utility metric under different evaluators yields a stable ranking compared to end-to-end LLM scoring. More details and examples of the validation study are provided in Appendix \ref{sec:human_eval}.


\textbf{Qualitative Evaluation:} To complement the quantitative validation, we examine specific outputs in detail, illustrating what high scoring paths look like in practice. Figure \ref{fig:case_study} shows examples that have quality at least 3, and which have high minimum distance to the population of paths from that query over all models. From the resulting scored pairs, we chose 3 representative samples.

First, \textbf{the paths that are high-scoring under our metrics represent appropriately-scored relations.} The first and last are classic strong connections through acting. The second is weaker, although St.~Louis is a less common origin for bands than many other cities. Second,  these connections are novel, where each of these instances were found by only a single model. We want to note that different models surface different interesting connections, especially for tail entities, and no single model captures most of them.


\begin{figure*}[t!]
\centering
\includegraphics[scale=0.25, trim= 0 24cm 0 0]{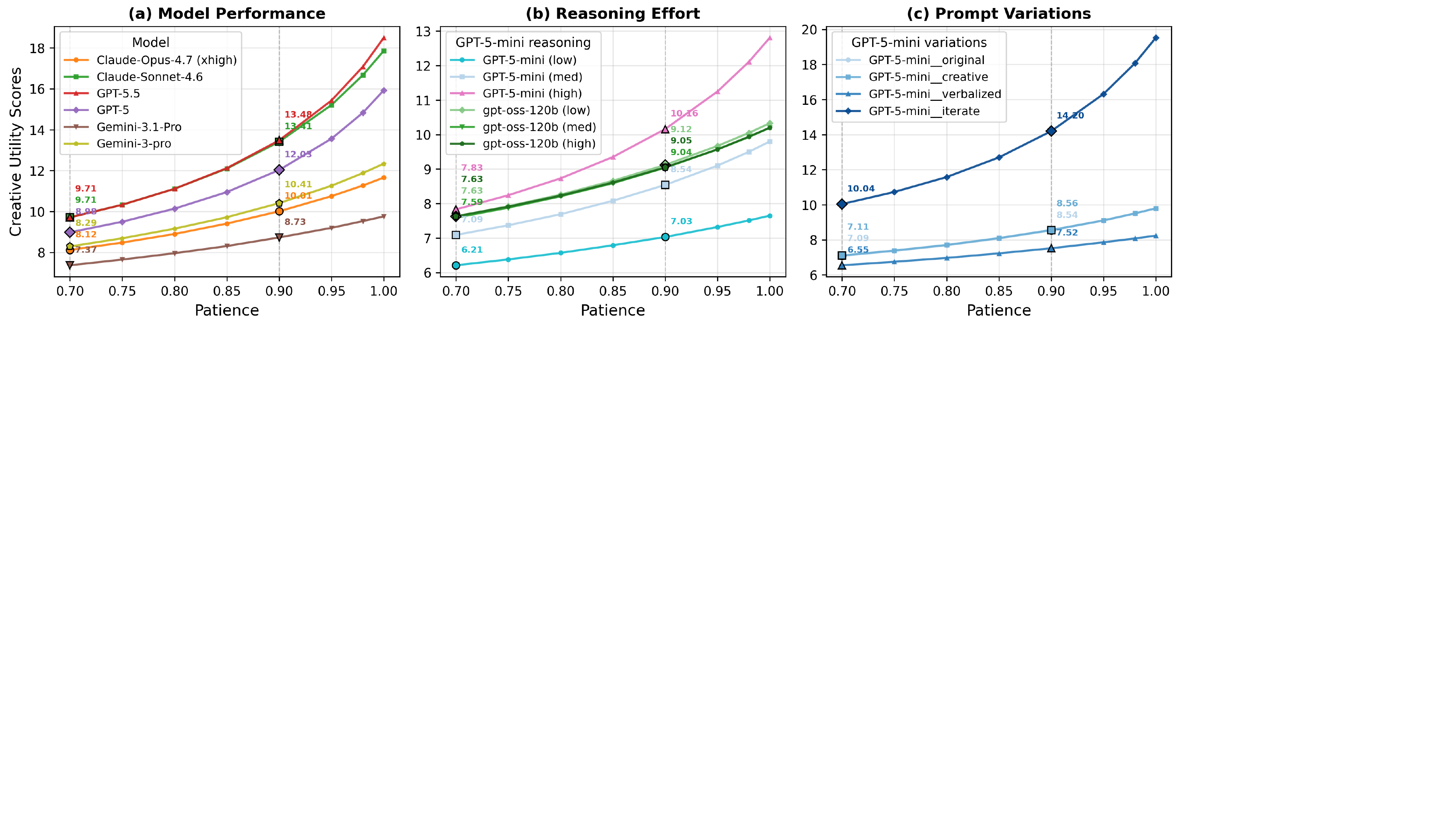}
\vspace{-1em}
\caption{Model performance on CREATE. We report the creative utility vs.~patience values for (a) different frontier models (b) different reasoning effort for GPT-5-mini and gpt-oss-120b (c) different prompt variations for GPT-5-mini. GPT-5.5 and Claude-Sonnet-4.6 achieve the best scores; we explain the low performance of some frontier models in the text.}
\vspace{-1em}
\label{fig:results_1}
\end{figure*}

\section{Results}  \label{sec:rq1}

\textbf{Frontier models achieve the highest creative utility.} Figure \ref{fig:results_1} reports creative utility of our different target LLMs on our benchmark at a range of patience values. With patience 1 (no decay in contribution for subsequent paths, except as given by the distance function), LLMs score as high as 18, indicating that at least 18 paths are found on average. GPT-5.5, Claude-Sonnet-4.6, and GPT-5 are ranked first, second, and third across patience values. 
Moreover, differences between models become more pronounced as patience increases, highlighting how models sustain quality and diversity over longer sets of generations. At patience 1, the gaps between the top two models (GPT-5.5/Claude-Sonnet-4.6) and the rest are statistically significant ($p<0.05$); see Appendix~\ref{sec:stats_testing} for details.

\textbf{Certain models are conservative on this task.} A few models notably underperform on the task. Gemini 3.1-Pro performs the worst, partially because it is hesitant to generate answers that could be non-factual, despite being asked to in the prompt. Claude-Opus-4.7 also performs poorly even at high reasoning effort. We attribute this to the adaptive thinking capability of this model; we do not see that it thinks long enough on these problems to produce a substantial number of paths. 


\textbf{Reasoning has a mixed impact.} In the second subfigure of Figure \ref{fig:results_1}, we see that GPT-5-mini benefits substantially from a larger reasoning budget, while gpt-oss-120b shows little to no improvement. This suggests that the returns to additional reasoning are model-specific rather than universal, and may depend on how effectively a model uses its reasoning trace to plan diverse, high-utility paths.

In Table \ref{tab:combined_summary_table1} we report average maximum quality of model generated paths, average pairwise distance within a set, number of factual paths generated, and length of paths generated. These values additionally allow us to conclude that while utility increases with number of paths, some models do well with low path counts: for instance, Gemini-3-Pro obtains similar utility to GPT-5-mini (high reasoning) despite producing fewer paths, by having higher quality and more distinct paths. 
Appendix ~\ref{app:results} shows examples of raw predictions from GPT-5, Gemini-3-Pro, and Claude-Haiku-4.5, along with examples of paths with high and low quality quality scores (before factuality filtering).

\textbf{Prompt-based interventions weakly and inconsistently affect creativity across models.} In Figure \ref{fig:results_1} (right side), we show how the utility values change with prompting variations for GPT-5-mini. Iterative prompting is the most effective in increasing the utility, which is expected since these approaches introduce additional generation paths that directly contribute to the metric.

\begin{wrapfigure}{r}{0.45\textwidth}
\centering
 \includegraphics[scale=0.3, trim= 0 27cm 6cm 0]{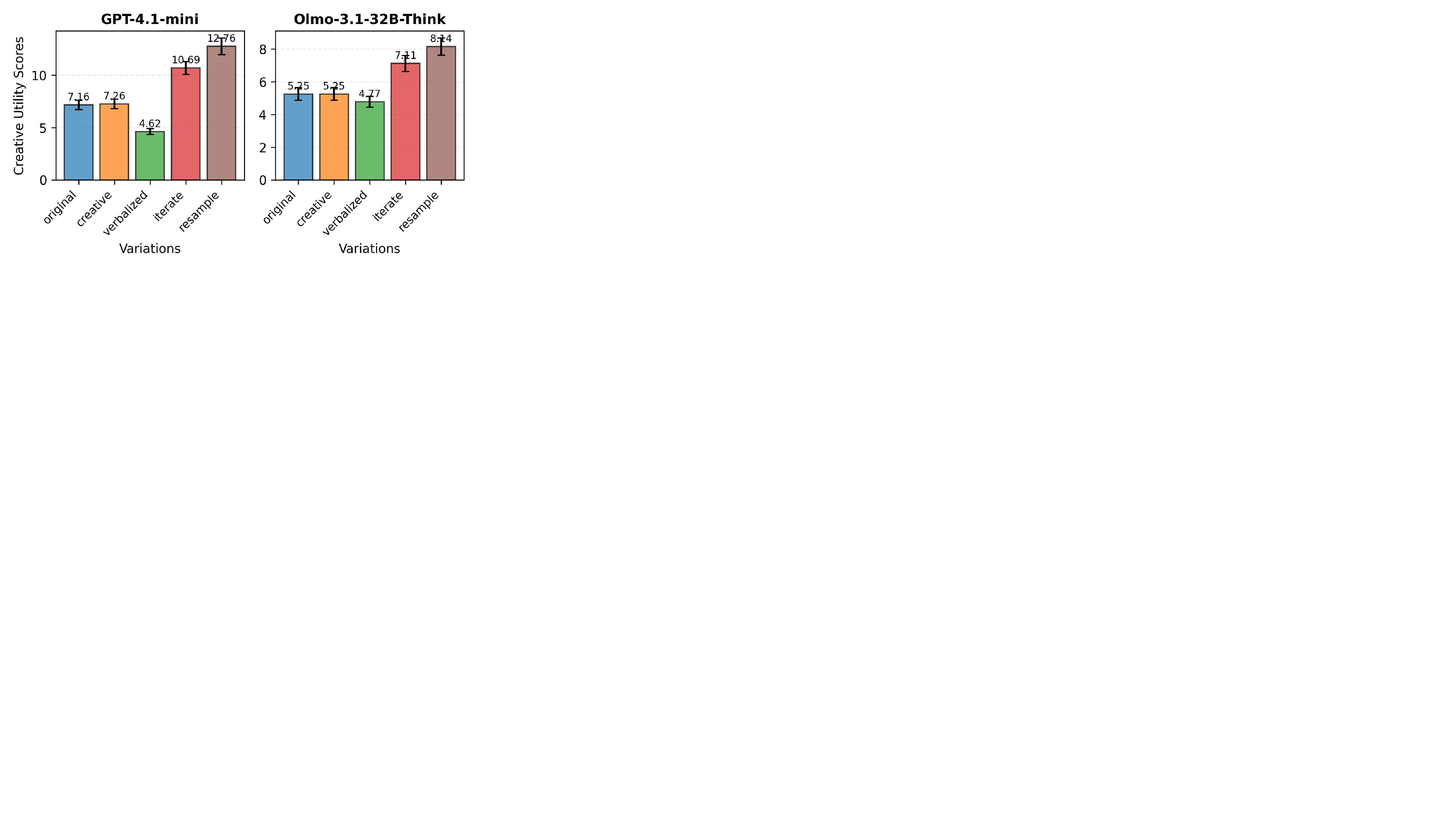}
\vspace{-1em}
\caption{Alternative prompting methods can lead to improvements depending on the model. Iterate and Resample interventions lead to the highest creative utility scores.}
\label{fig:pv_variations}
\end{wrapfigure}

Figure \ref{fig:pv_variations} illustrates how utility changes for different models. For prompt variations, `creative' and `original' prompts perform very similarly across models; they also tend to produce similar number of paths post validity and factuality filtering. `verbalized' prompting produces very low number of valid paths and factual paths. Despite the strength of iterative prompting, resampling performs the best; however, both of these techniques leverage a much larger inference budget on the task. Detailed results for this experiment are given in Table~\ref{tab:table2_variations}.

\textbf{Models produce similar paths overall.} We measure how a generated path differs from the population of paths generated by all models for the same query. We focus on high-quality responses (quality > 3) across 300 queries. For every path that satisfies the criteria above, we calculate each path's distance from a population of responses for the same query. Table \ref{tab:high_distinct} reports the average distance from each path to its closest neighbor in the population. Values are largely similar across models and prompt variations, with ``iterate'' standing out as the most effective strategy, suggesting that explicitly instructing a model to diverge from its prior outputs yields more distinct responses than resampling alone. Despite the small average distances, there is a small fraction of paths with high distance values from the population that are creatively interesting, as we show in Section \ref{sec:case_study}. See Appendix \ref{app:distinctiveness} for full methodology and the results per model.


\begin{table}[]
\centering
\small
\caption{Examples of snippets from reasoning traces of open-source models. LLMs show different problem solving behavior employing different strategies to brainstorm connections }
\label{tab:reasoning_analysis}
\begin{tabular}{l p{0.8\linewidth}}
\toprule
Strategy & Excerpt \\
\midrule
Goal Setting & \textbf{Qwen3-32B}: \textit{First, I need to start with David Koechner, the actor. Target is someone whose place of birth is Newport Beach, paths need to end with that triple.} \\
\makecell[l]{Feature \\ Enumeration} & \textbf{Qwen3-32B}: \textit{First, the obvious paths might be through his acting roles. He's in `Anchorman,' `The Hangover,' etc. But maybe someone he worked with was born there. Like, maybe a co-star in one of his movies is from Newport Beach. … For example, in `Anchorman,' the main cast includes Will Ferrell, Steve Carell, etc.} \\

Backward Chaining & \textbf{OLMo-3.1-32B-Think }: \textit{I think I have to make do with what I can think of. Let me try to outline possible paths:
1. David Koechner is a co-star with [X], who was born in Newport Beach. (Need to find X)} \\
\makecell[l]{Bidirectional \\Enumeration} & \textbf{gpt-oss-120b}: \textit{Let's think of people born in Newport Beach, CA: Notable: Michael Clarke Duncan? No, he was born in Chicago. Let's list: Nicole Kidman? No. Some celebrities: Beau Bridges? Born in Los Angeles.} \\
\bottomrule
\end{tabular}
\end{table}


\section{Reasoning Trace Analysis}

Table \ref{tab:reasoning_analysis} presents excerpts from the reasoning traces of several open-source models as they attempt to solve CREATE queries. These excerpts illustrate a handful of problem-solving strategies we manually identified. Among the strategies shown, we observe \textit{goal setting}, in which the model establishes a direction and lists the task constraints, and \textit{feature enumeration}, in which the model systematically iterates through known properties of the head entity. 
We also find evidence of \textit{backward chaining} where the model first sketches outlines of candidate solution paths and then progressively fills in the missing details. Finally, we observe \textit{bidirectional enumeration}, where the model tends to brainstorm connections from both ends of the query, reasoning from the head entity while simultaneously enumerating from the tail entity. Figure \ref{fig:trace_example} shows a reasoning trace with strategies highlighted, including enumeration (yellow and orange text) and backward chaining (green text). One thing to note from the reasoning chain is that models tend to revisit the same nodes leading to some repeated information, which undermines the effectiveness of the search process. Characterizing these strategies and their effectiveness reveals where reasoning helps versus where it hurts. This points to concrete targets for improving search efficiency and informs the design of training signals and inference-time interventions for better LLM creativity.




\section{Conclusion} 
This paper presents a benchmark for testing LLMs at associative creativity. Queries require models to find a set of strong, distinct connections between real concepts, mimicking the kind of associative creativity required more broadly, but maintaining evaluatability. Our results highlight strengths of current models, but also show that higher thinking effort is not necessarily a silver bullet in these settings. Further work is needed to fully leverage LLMs for tasks requiring associative creativity.

\section*{Acknowledgments}
This work was partially supported by NSF CAREER Awards IIS-2145280 and IIS-2145479, NSF grants IIS-2433071 and IIS-2107524, NIH grant 1R01LM01460001, the NSF AI Institute for Foundations of Machine Learning (IFML), and the NSF-Simons AI Institute for Cosmic Origins (CosmicAI) under NSF Cooperative Agreement 2421782 and Simons Foundation MPS-AI-00010515. This work is also partially supported by the Sloan Foundation and grants from Amazon and Open Philanthropy. This research has been supported by computing support from the Torch cluster at NYU as well as a compute grant from NVIDIA. 

\bibliographystyle{plain}
\bibliography{neurips_2026.bib}

\appendix

\section{Limitations}

This dataset is limited to the English language and is based on facts available on Wikidata, although not restricted to those facts at evaluation time. It reflects a curated set of domains (people, genes and medical drugs) meant to be representative of trivia connections and ideation, but without perfect coverage of all real-world applications. As a result, this dataset may not perfectly reflect associative creativity in other languages or other domains; however, the metric that we propose is generally applicable, and the approach to constructing the dataset is transferable to other domains as well.

\section{Benchmark} \label{app:benchmark}

We describe the data generation process in Algorithm \ref{alg:nlq-generation}. To ensure quality and non-trivial connections, we remove any links where the source entity also satisfies the tail connection. We also remove any relations $(r,c)$ where $r$ induces a class that has a very large membership (e.g. \textit{native language, instance of, date of birth, gender, described at URL} etc). We show examples of queries generated by our system in Table \ref{tab:benchmark}. Statistics about the relations used in our benchmark along with the number of entities and queries are given in Table ~\ref{tab:relation_stats}. A source path generated according to our benchmark is about three triples long.

\begin{algorithm}
\small
\caption{Generate Natural Language Queries from a Knowledge Graph}
\label{alg:nlq-generation}
\begin{algorithmic}[1]
\REQUIRE Knowledge graph $G=(E,R)$; category set $C=\{(r,c)\mid r\in R,\,c\in E\}$; text model $\texttt{M}(\cdot)$
\ENSURE Query list $Q$
\STATE $Q \leftarrow \emptyset$ \COMMENT{collected natural-language queries}
\FORALL{$(r,c)\in C$}
    \STATE $C_{r,c} \leftarrow \{x\in E \mid (x,r,c)\in G\}$ \COMMENT{entities in class defined by $(r,c)$}
    \STATE $\mathcal{P} \leftarrow \{\{x_1,x_2\}\subseteq C_{r,c}\mid x_1\neq x_2\}$ \COMMENT{unordered entity pairs within the class}
    \FORALL{$\{x_1,x_2\}\in \mathcal{P}$}
        \STATE sample uniformly $x \in \{x_1,x_2\}$ \COMMENT{choose one endpoint to expand by one hop}
        \STATE $\mathcal{O}(x) \leftarrow \{(x,r',y)\in G \mid r'\in R,\,y\in E\}$ \COMMENT{outgoing edges from $x$}
        \STATE sample uniformly $(x,r',y)\in \mathcal{O}(x)$ \COMMENT{pick an informative 1-hop fact}
        \STATE $s \leftarrow \big((x_1,r,c),\,(x_2,r,c),\,(x,r',y)\big)$ \COMMENT{source path: shared class + extra hop}
        \STATE $q \leftarrow \texttt{M}\!\left(\text{prompt}\big([x_1,\_,\_],\,[\_,r',y]\big)\right)$ \COMMENT{convert $s$ into NL query}
        \STATE $Q \leftarrow Q \cup \{q\}$ \COMMENT{store query for evaluation}
    \ENDFOR
\ENDFOR
\STATE \textbf{return} $Q$
\end{algorithmic}
\end{algorithm}

\newcolumntype{Y}{>{\RaggedRight\arraybackslash}X}
\begin{table*}[h]
\centering
\small
\setlength{\tabcolsep}{4pt}
\renewcommand{\arraystretch}{1.15}

\caption{Examples of instances of (relation, category) in our dataset, along with pairs of triples retrieved that satisfy this category, expanded triple 2, and examples of the constructed query.}
\label{tab:benchmark}
\begin{tabularx}{\textwidth}{@{} l Y Y Y @{}}
\toprule
(Relation, Category) & Triples retrieved & Triple 2 expanded & Example Query \\
\midrule

\multirow{2}{*}{(cast member, Goodfellas)} &
\textbf{Triple 1:} (Goodfellas, cast member, Robbie Vinton) &
\multirow{2}{=}{(Vincent Gallo, occupation, painter)} &
\multirow{2}{=}{What are different ways of connecting Robbie Vinton, an American Actor, and someone who is a painter?} \\
&
\textbf{Triple 2:} (Goodfellas, cast member, Vincent Gallo) & & \\
\addlinespace

\midrule 

\addlinespace
\multirow{2}{*}{(member of sports team, Scuderia Toro Rosso)} &
\textbf{Triple 1:} (Sebastian Vettel, member of sports team, Scuderia Toro Rosso) &
\multirow{2}{=}{(Max Verstappen, unmarried partner, Kelly Piquet)} &
\multirow{2}{=}{What are different ways of connecting Sebastian Vettel, the German racing driver, and someone who has been/is a unmarried partner of Kelly Piquet?} \\
&
\textbf{Triple 2:} (Max Verstappen, member of sports team, Scuderia Toro Rosso) & & \\
\addlinespace
\midrule 
\addlinespace

\multirow{2}{*}{(medical condition treated, ulcerative colitis)} &
\textbf{Triple 1:} (prednisolone, medical condition treated, ulcerative colitis) &
\multirow{2}{=}{(methotrexate, subject has role, folic acid antagonists)} &
\multirow{2}{=}{What are different ways of connecting prednisolone, a medication used to treat various conditions, and a substance that plays a role as a folic acid antagonist?} \\
&
\textbf{Triple 2:} (methotrexate, medical condition treated, ulcerative colitis) & & \\
\addlinespace
\addlinespace
\addlinespace
\bottomrule
\end{tabularx}
\end{table*}

\paragraph{Benchmark License} Our benchmark is licensed under CC BY-SA 4.0.

\section{Distance Function} \label{app:distance}
In Section ~\ref{subsec:formal}, we introduce the distance function used for calculating the creative utility metric. Table ~\ref{tab:distance_examples} shows examples of pairs of paths, along with the raw cosine distance and the transformed values. Paths with values cosine distance $>$0.7 are substantially different, focusing on different domains and properties of the entities (for example, familial connections vs professional service), where as those under 0.4 have similarities in domains of properties traversed, entities and sometimes are paraphrases of each other (for example, connections via position in office or both paths focusing on  familial connections). 

\begin{wraptable}{r}{0.50\textwidth}  
\vspace{-1.0\baselineskip} 
\small
\renewcommand{\tabcolsep}{1.4mm}
\caption{List of source relations, number of unique entities, number of queries along with the average \# of source paths per relation.}
\label{tab:relation_stats}
\begin{tabular}{l l c c}
\hline
 ID &  Label & Unique $e_1$ & \# Queries \\
\hline
P166 & award received & 99 & 99 \\
P39 & position held & 93 & 94 \\
P40 & child & 86 & 88 \\
P463 & member of & 88 & 89 \\
P54 & member of sports team & 99 & 99 \\
P57 & director & 30 & 47 \\
P58 & screenwriter & 48 & 60 \\
P161 & cast member & 39 & 96 \\
P162 & producer & 43 & 65 \\
P2175 & medical condition treated & 96 & 100 \\
P2293 & genetic association & 94 & 94 \\
\hline
\end{tabular}
\end{wraptable}

\begin{figure}[t]
\centering
\includegraphics[scale=0.28,trim=0 26cm 0 0cm]{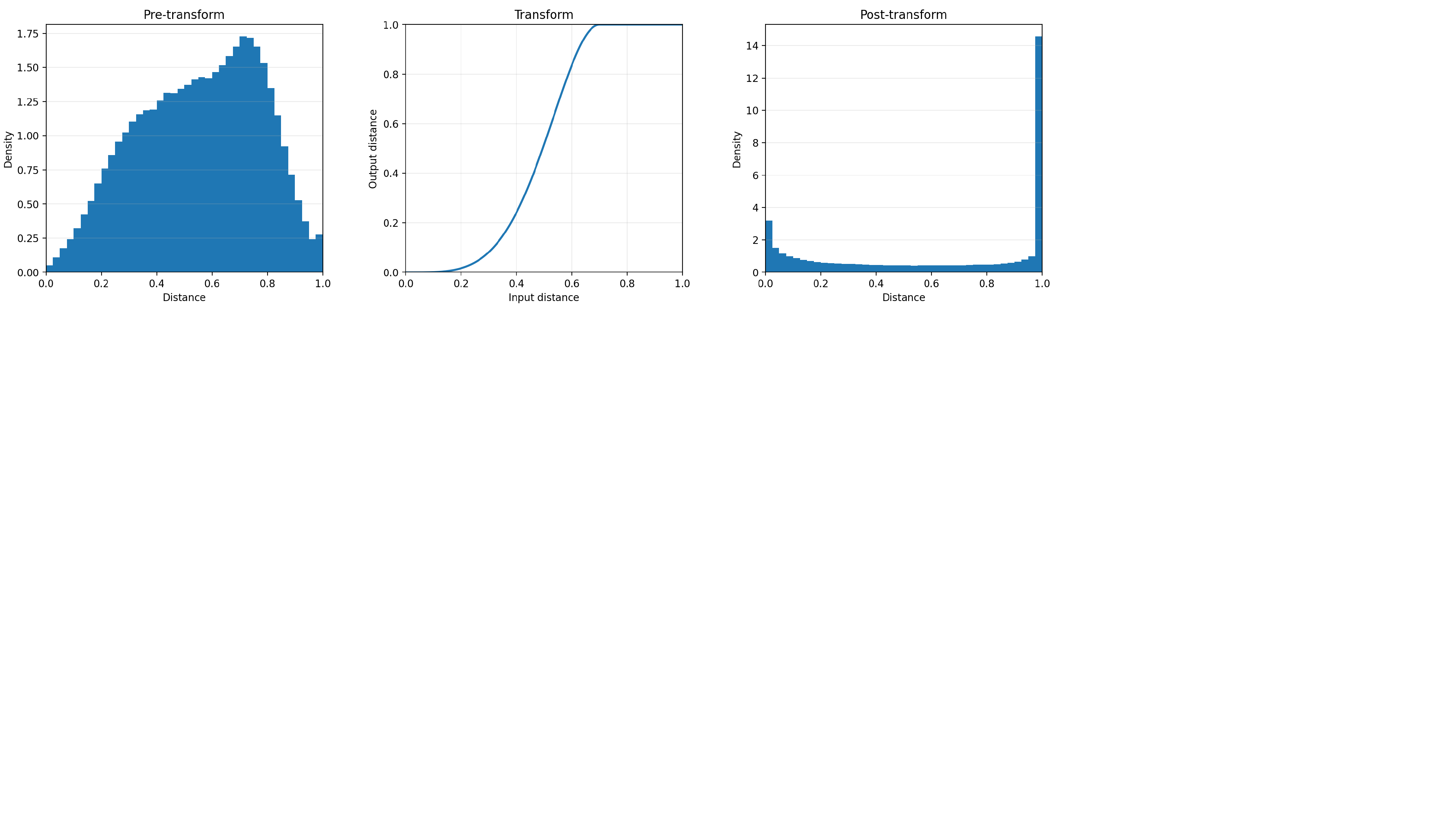}
    \caption{Pre-transformation, transformation function, and post-transformation cosine distance scores for GPT-5 (medium) paths. Many pairs being scored at a distance of 1 lined up with our intuitive assessment that they represented completely different relationships.}
    \label{fig:distance_transformation}
\end{figure}

\begin{table}
\centering
\small
\caption{Examples of paths, raw distance values, and distance values post transformation}
\label{tab:distance_examples}
\setlength{\tabcolsep}{4pt}
\begin{tabular}{p{0.38\linewidth} p{0.38\linewidth} c c}
\toprule
Path 1 & Path 2 & \makecell{cosine \\ distance} & $g(x)$ \\
\midrule
((`george h. w. bush', `successor as president to', `ronald reagan'), (`ronald reagan', `spouse', `nancy reagan'), (`nancy reagan', `position held', `first lady of the united states')) & ((`george h. w. bush', `vice president to', `ronald reagan'), (`ronald reagan', `spouse', `nancy reagan'), (`nancy reagan', `position held', `first lady of the united states')) & 0.13 & 0.003 \\
\\
((`lee jong wook', `was director-general of', `world health organization'), (`world health organization', `has employee educated at', `london school of hygiene \& tropical medicine')) & ((`lee jong wook', `worked at', `world health organization'), (`world health organization', `employed', `david nabarro'), (`david nabarro', `educated at', `london school of hygiene \& tropical medicine')) & 0.33 & 0.12 \\
\\
((`josefa edralin marcos', `relative', `michael marcos keon'), (`michael marcos keon', `occupation', `lawyer')) & ((`josefa edralin marcos', `spouse', `ferdinand marcos'), (`ferdinand marcos', `legal mentor of', `benjamin abalos jr.'), (`benjamin abalos jr.', `occupation', `lawyer')) & 0.40 & 0.24 \\
\\
((`christopher nolan', `sibling', `jonathan nolan'), (`jonathan nolan', `spouse', `lisa joy')) & ((`christopher nolan', `directed', `the dark knight trilogy'), (`the dark knight trilogy', `screenplay by', `jonathan nolan'), (`jonathan nolan', `spouse', `lisa joy')) & 0.58 & 0.77 \\
\\
((`joe spinell', `worked with director', `martin scorsese'), (`martin scorsese', `place of birth', `new york city'), (`new york city', `includes borough', `brooklyn')) & ((`joe spinell', `acted in tv movie', `she waits'), (`she waits', `produced by', `abc'), (`abc', `headquarters in', `new york city'), (`new york city', `includes borough', `brooklyn')) & 0.71 & 1.0 \\
\\
((`george h. w. bush', `associated with', `bush family'), (`bush family', `includes', `barbara bush'), (`barbara bush', `position held', `first lady of the united states')) & ((`george h. w. bush', `served as', `director of central intelligence'), (`director of central intelligence', `advised', `hillary clinton'), (`hillary clinton', `position held', `first lady of the united states')) & 0.8 & 1.0 \\
\bottomrule
\end{tabular}
\end{table}

\section{Evaluation Prompts}  \label{sec:eval_prompts}
Prompts \ref{prompt:strength_prompt} and \ref{prompt:factuality_prompt} are used for evaluating specificity and factuality.

For the specificity prompt, we elicit class sizes per triple from the model. Once we have the class sizes we use the following mapping to convert the size to a 1-5 scale.

$q(\text{class\_size}) =
\begin{cases}
5, & \text{if } \text{class\_size} \le 10, \\
4, & \text{if } 10 < \text{class\_size} < 100, \\
3, & \text{if } 100 \le \text{class\_size} < 500, \\
2, & \text{if } 500 \le \text{class\_size} < 5000, \\
1, & \text{otherwise}
\end{cases}$.

\section{Metric Validation} \label{sec:human_eval}

\begin{wraptable}{r}{0.4\textwidth}
    \small
    \centering
    \renewcommand{\tabcolsep}{0.55mm}
    \caption{LLM-as-a-judge performance for evaluating factuality of a generated path.}
    \label{tab:classification_report}
    \begin{tabular}{lrrrr}
    \toprule
    Class & Precision & Recall & F1-Score & Support \\
    \midrule
    0 & 0.52 & 0.94 & 0.67 & 72 \\
    1  & 0.98 & 0.77 & 0.87 & 274 \\
    \bottomrule
    \end{tabular}
\end{wraptable}

\subsection{Factuality}
\label{sec:factuality_human_eval}

In this section, we report the detailed performance metrics for {\small\texttt{gpt-oss-120b}} as the evaluator for factuality. We annotated 346 relations (triples) and then validated the evaluator against our judgments. We report the precision/recall for the binary classification in Table~\ref{tab:classification_report}. The balanced accuracy for the evaluator is 0.86. 

The most common sources of errors for this factuality evaluator are, firstly, the model incorrectly identifies and assesses generations involving niche and long tail entities, which points to a knowledge gap. Secondly, the model often misinterprets relations and entities leading to false judgments. Often if the entity is too general, it is evaluated strictly, even though the relation holds true between the general entities.  Table ~\ref{tab:factuality_examples} shows two examples of LLM judgment on triples generated by one of models.

\begin{wraptable}{r}{0.4\textwidth}
\small
\centering
\renewcommand{\tabcolsep}{0.55mm}
\caption{Performance of different models as the factuality evaluator. We report the binary accuracy, the F1 for the positive i.e. factual class (pos) and the negative i.e. hallucinated class (neg).}
\label{tab:factuality_diff_models}
\begin{tabular}{lrrr}
\toprule
Model & Binary Acc & F1(pos) & F1 (neg) \\
\midrule
gpt-oss-120b & \textbf{0.86} & \textbf{0.87} & \textbf{0.67} \\
GPT-5-mini & 0.82 & 0.87 & 0.67 \\
GPT-4o-mini & 0.69 & 0.83 & 0.53 \\
GPT-4.1-mini & 0.68 & 0.85 & 0.50 \\
\bottomrule
\end{tabular}
\end{wraptable}

\begin{table}[]
    \centering
    \small 
    \caption{Factuality Judgments and Explanations for Knowledge Triples. The judgments highlighted are incorrect because of entities and the relation being too long tail.}
    \label{tab:factuality_examples}
    \begin{tabularx}{\textwidth}{@{}p{4cm} X p{2.5cm}@{}}
        \toprule
        Triples & LLM Explanation & LLM Judgments \\
        \midrule
        
        (rick mcphail, associated musical artist, thees uhlmann) \newline 
        (thees uhlmann,country of citizenship,germany)
        & 
        The first triple links `rick mcphail' to `thees uhlmann' as an associated musical artist. While Thees Uhlmann is a real German musician, there is no verifiable record of a musician named Rick McPhail, nor any known collaboration or association between them. The lack of evidence for the existence of Rick McPhail as a musical artist makes this triple hallucinated. The second triple states that Thees Uhlmann's country of citizenship is Germany. This matches publicly available biographical information about Uhlmann, confirming the claim as accurate and not hallucinated. 
        & 
        \textbf{hallucinated}, not hallucinated \\
        \midrule

        (robert metcalfe, educated at, harvard university) \newline 
        (harvard university, employer, edmund m. clarke] \newline 
        (edmund m. clarke, doctoral student, sergio vale aguiar campos)
        & 
        The first triple is correct: Robert Metcalfe earned his Ph.D. from Harvard University, so both entities exist and the 'educated at' relation is factual. The second triple claims Harvard University employs Edmund M. Clarke; Clarke was a long‑time professor at Carnegie Mellon University and has no known employment by Harvard, making this claim unsupported. The third triple asserts that Edmund M. Clarke supervised a doctoral student named Sergio Vale Aguiar Campos; there is no public record of such a student under Clarke, suggesting the entity or relationship is fabricated. Therefore, only the first triple is not hallucinated. 
        & 
        not hallucinated, \textbf{hallucinated}, \textbf{hallucinated} \\
       
        \bottomrule
    \end{tabularx}
\end{table}
 
\subsection{Comparing Different LLM-as-a-judge models}

To evaluate the choice of model for the LLM-as-a-judge, we compare different LLMs against our human annotations. We test GPT-5-mini, GPT-4o-mini and GPT-4.1-mini. From Tables \ref{tab:factuality_diff_models} and \ref{tab:specificity_diff_models} we see that our choice of the model i.e. gpt-oss-120b does fairly well given the cost and the effectiveness.

\subsection{Human annotations on creativity}

We give the following instructions to human annotators for evaluating the creativity/interestingness of the model generated outputs. The annotations are on a scale of 1-5. 

\textbf{1: Not creative}:
Nearly all paths pass through the same intermediate concept(s) or domain. Variations are superficial, different relation labels but the same nodes. The set feels like one idea restated.

\textbf{2: Slightly creative}:
Most paths cluster around one dominant intermediate concept or domain, with one or two outliers. The outliers are either weak (obscure intermediate nodes) or structurally similar to the rest. Minimal semantic spread.

\textbf{3: Moderately creative}:
Paths span distinct intermediate domains. Some clustering remains but there is genuine topical breadth. Intermediate nodes are reasonably well-known. Few paths are meaningfully different from each other.

\textbf{4: Creative}:
Paths traverse meaningfully distinct intermediate domains via salient, well-known entities. Topical clustering is limited. The core set of paths is diverse — some paths add new angles rather than pad existing ones.

\textbf{5: Highly creative}:
Paths spread across a wide semantic space. Almost no two paths share a central intermediate concept. Intermediate entities are salient and important in their respective domains. The set feels like a genuine, broad-coverage map of how the two entities can be meaningfully linked.

\begin{wraptable}{r}{0.6\textwidth}
\small
\centering
\renewcommand{\tabcolsep}{0.55mm}
\caption{Pearson's correlation of specificity judgments between humans and different models}
\label{tab:specificity_diff_models}
\begin{tabular}{l|rrrr}
\toprule
Ann & gpt-oss-120b & GPT-4o-mini & gpt-41-mini & GPT-5-mini \\
\midrule
A & 0.64 & 0.78 & 0.56 & 0.81 \\
B & 0.58 & 0.63 & 0.47 & 0.75 \\
avg & 0.67 & 0.78 & 0.56 & 0.85 \\
\bottomrule
\end{tabular}
\end{wraptable}


\section{Stability of Creative Utility vs.~Stability of End-to-End Evaluation} \label{subsec:stability}

We compare two approaches for evaluating outputs on CREATE: creative utility and end-to-end evaluation of creativity with LLM-as-a-judge. We show in this section that creative utility is more stable across LLMs: due to being grounded in objective quantities like class sizes, results vary less as the judge model is changed.

To evaluate this, we re-ran our evaluations using different LLMs. 
We sampled a subset of 165 instances (15 per relation type) with outputs from the following models: GPT-5, Gemini-3-pro, Claude-Haiku-4.5-med, and GPT-5-mini (at high, medium, and low reasoning levels). In addition to our standard gpt-oss-120b evaluator, we explored two additional models: GPT-4o-mini and GPT-4.1-mini.

\begin{table}[]
\small
\centering
\renewcommand{\tabcolsep}{0.7mm}
\caption{This table shows (a) the stability of our proposed metric under different evaluators (b) how our proposed creative utility metric is more stable compared to simply prompting an LLM for creativity judgments.}
\label{tab:stability}
\begin{tabular}{l|rr|rr|rr}
\toprule
\multicolumn{7}{c}{Proposed Creative Utility} \\
\midrule
\multicolumn{1}{c}{model} & \multicolumn{1}{c}{GPT-4.1-mini} & \multicolumn{1}{c}{rank} & \multicolumn{1}{c}{GPT-4o-mini} & \multicolumn{1}{c}{rank} & \multicolumn{1}{c}{gpt-oss-120b} & \multicolumn{1}{c}{rank} \\
\midrule
GPT-5-mini (low) & 9.94 & 5 & 7.86 & 5 & 6.81 & 5 \\
GPT-5-mini (medium) & 11.30 & 4 & 8.99 & 4 & 8.58 & 4 \\
GPT-5-mini (high) & 12.58 & 3 & 10.23 & 3 & 10.19 & 3 \\
GPT-5 & 13.41 & 1 & 11.33 & 1 & 12.34 & 1 \\
Claude-Haiku-4.5 & 7.48 & 6 & 6.79 & 6 & 5.54 & 6 \\
Gemini-3-Pro & 13.07 & 2 & 10.91 & 2 & 10.68 & 2 \\
\midrule
\multicolumn{7}{c}{Single prompt w/ LLM as a judge} \\
\midrule
GPT-5-mini (low) & 3.63 & 3 & 4.29 & 5 & 3.07 & 5 \\
GPT-5-mini (medium) & 3.58 & 4 & 4.33 & 3 & 3.13 & 4 \\
GPT-5-mini (high) & 3.65 & 2 & 4.41 & 2 & 3.25 & 2 \\
GPT-5 & 3.51 & 5 & 4.32 & 4 & 3.16 & 3 \\
Claude-Haiku-4.5 & 3.05 & 6 & 3.84 & 6 & 2.63 & 6 \\
Gemini-3-Pro & 3.99 & 1 & 4.53 & 1 & 3.58 & 1 \\ 
\bottomrule
\end{tabular}
\end{table}

As shown in Table \ref{tab:stability}, our creative utility metric remains stable across different LLMs used for factuality and specificity evaluation. It is notably more stable than prompting an LLM directly for creative utility judgments, where scores vary by over an entire point on a 1-5 scale.
To quantify this further, we computed Krippendorff's alpha between the three evaluators across all six models at the instance level (~900 instances total). On our creative utility metric, our LLM annotators achieve an alpha of 0.76, compared to just 0.31 for direct prompting. This substantial gap indicates that our decomposed metric yields far more consistent rankings across evaluator models, whereas direct rubric-based prompting is highly sensitive to the choice of LLM judge.

\section{Qualitative Validation}

\begin{table}[t] 
\centering
\small
\caption{Examples of model-generated paths $u$, the population path $u$ is closest to, along with quality scores and minimum distance values.}
\label{tab:case_study_app}
\setlength{\tabcolsep}{4pt}
\begin{tabular}{p{0.43\linewidth} c p{0.43\linewidth} c}
\toprule
Path & $\sigma$ & population path & $d$ \\
\midrule
(jean-pierre jeunet, directed, micmacs), (micmacs, cast member, yolande moreau), (yolande moreau, voice actor in, mia and the migoo), (mia and the migoo, nominated for, european film award for best animated feature film) & 4 & (jean-pierre jeunet, directed, micmacs), (micmacs, featured animation by, annecy international animated film festival winners), (annecy international animated film festival winners, nominated for, european film award for best animated feature film) & 0.40 \\
\midrule
(jean-pierre jeunet, directed, delicatessen), (delicatessen, cast member, karin viard), (karin viard, voice actor in, the suicide shop), (the suicide shop, nominated for, european film award for best animated feature film) & 4 & (jean-pierre jeunet, directed, delicatessen), (delicatessen, nominated for, european film award for best animated feature film) & 0.18 \\
\midrule
(jean-pierre jeunet, directed, amélie), (amélie, cast member, jamel debbouze), (jamel debbouze, voice actor in, why i did (not) eat my father), (why i did (not) eat my father, nominated for, european film award for best animated feature film) & 4 & (jean-pierre jeunet, directed, amélie), (amélie, features, jamel debbouze), (jamel debbouze, featured in, a monster in paris), (a monster in paris, nominated for, european film award for best animated feature film) & 0.04 \\
\midrule
(jean-pierre jeunet, directed, amélie), (amélie, cast member, mathieu kassovitz), (mathieu kassovitz, voice actor in, april and the extraordinary world), (april and the extraordinary world, nominated for, european film award for best animated feature film) & 4 & (jean-pierre jeunet, directed, amélie), (amélie, starred, mathieu kassovitz), (mathieu kassovitz, voice actor in, april and the extraordinary world), (april and the extraordinary world, nominated for, european film award for best animated feature film) & 0.00  \\ 
\bottomrule
\end{tabular}
\end{table}

In Table \ref{tab:case_study_app} we show all valid, factually correct paths with quality $\ge$3 for the query: \textit{`What are different ways of connecting Jean-Pierre Jeunet, the French film director, and someone who was nominated for the European Film Award for Best Animated Feature Film?'}. For our analysis in Section \ref{sec:case_study}, from a set of predictions for this query, we pick the path with the highest minimum distance from the population. This path is the most \textit{distinctive} in the set of paths generated compared to a population. If we look at some paths with a lower distance score, we see that there is more overlap in the entities and relations explored, thereby showing that our metric is able to capture distinctive paths generated by the model.

\section{Inference Prompts and Variations} \label{app:inference_prompts}
Our inference pipeline has one base prompt and we also evaluate responses to different prompt variations to see if prompting has any influence on the creativity of responses. The base prompt is Prompt ~\ref{prompt:base_prompt}, the verbalized sampling prompt is Prompt ~\ref{prompt:verbalized_sampling}. For the `\textit{creative}' prompt variation, we simply append the instruction \textit{`- Be creative in the type of relationships explored and generated'} to the base prompt.  For the iterative prompt variation, the prompt is Prompt ~\ref{prompt:iterative}.

\section{Model Details} \label{app:model_details}

All inference was performed through LiteLLM, with proprietary models accessed via their respective APIs and open-source models served through self-hosted vLLM endpoints. For open-source models, we run the model inference using NVIDIA H200s GPUs. 

For open-source models we match the compute where possible, running inference at 16k and 32k tokens. For gpt-oss-120b we run inference at low/medium/high (as mentioned in the model specification). For thinking models, we stick to default or medium reasoning, with two exceptions: (a) Claude Opus, where we had to use high/xhigh because medium reasoning did not trigger the thinking since this model uses adaptive thinking; and (b) GPT-5-mini, where we experiment with the range of reasoning efforts available i.e. low/med/high.

\begin{table}[t]
\small
\centering
\caption{LLM checkpoints}
\begin{tabular}{l|ll}
\toprule
\textbf{Model}& \textbf{Checkpoint} & \textbf{reasoning budget} \\
\midrule
Olmo-3.1-32B-Instruct & allenai/Olmo-3.1-32B-Instruct & - \\
Olmo-3.1-32B-Think & allenai/Olmo-3.1-32B-Think & 16k / 32k \\ 
Qwen3-30B-Instruct & Qwen/Qwen3-30B-A3B-Instruct-2507 & - \\  
Qwen3-32B & Qwen/Qwen3-32B & 16k/32k \\
gpt-oss-120b & openai/gpt-oss-120b & low/med/high \\ 
Claude-Haiku-4.5  & \texttt{claude-haiku-4-5-20251001} & (default) \\ 
Claude-Sonnet-4.6 & \texttt{claude-sonnet-4-6} & (default) \\ 
Claude-Opus-4.7 & \texttt{claude-opus-4-7} & (default-high / xhigh) \\ 
GPT-4.1 & \texttt{gpt-4.1-2025-04-14} & - \\
GPT-4.1-mini & \texttt{gpt-4.1-mini-2025-04-14} & - \\
GPT-5-mini & \texttt{gpt-5-mini-2025-08-07} & low/med/high \\ 
GPT-5 & \texttt{gpt-5-2025-08-07} & default-med \\ 
GPT-5.4 & \texttt{gpt-5.4-2026-03-05} & default - med\\
GPT-5.5 & \texttt{gpt-5.5-2026-04-23} & default - med \\
Gemini-3-Pro & \texttt{gemini-3-pro-preview} & medium \\ 
Gemini-3.1-Pro & \texttt{gemini-3.1-pro-preview} & medium\\ 
\bottomrule
\end{tabular}
\label{tab:checkpoints}
\end{table}



\section{Results} \label{app:results}

\begin{figure}
\centering
\includegraphics[scale=0.5]{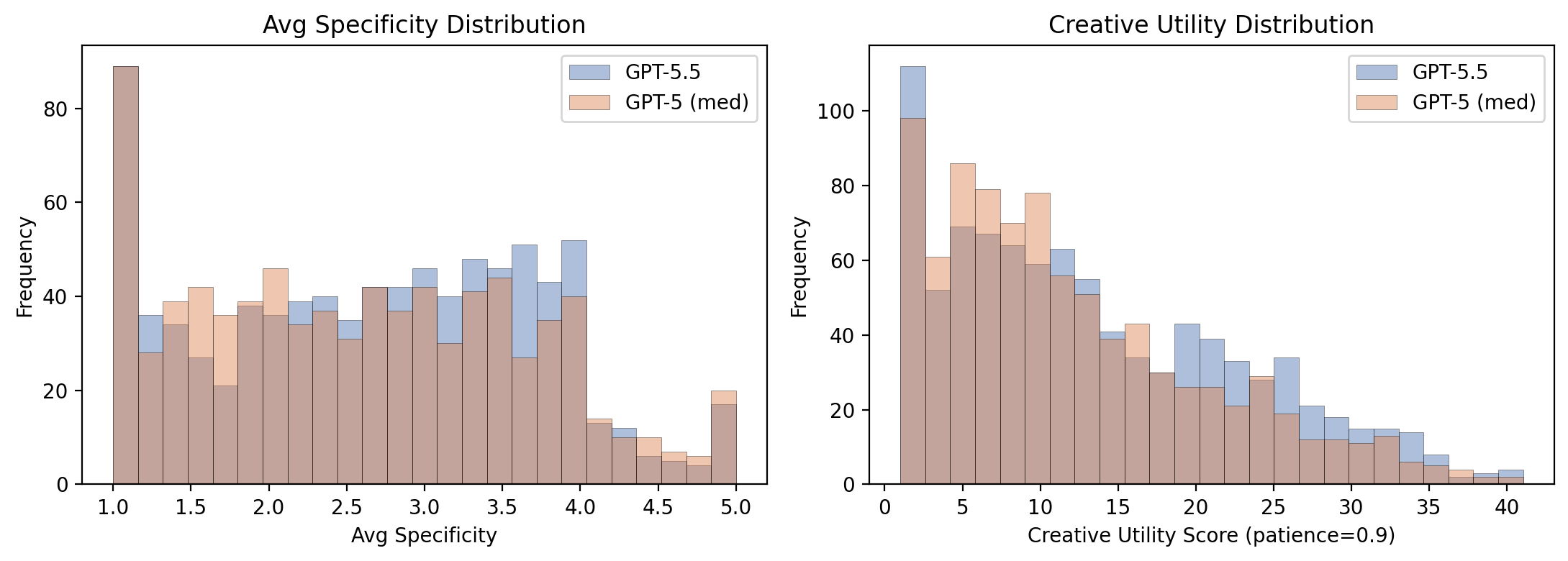}
    \caption{Distribution of creative utility scores for GPT-5.5 and GPT-5.}
    \label{fig:utility_distribution}
\end{figure}

Tables ~\ref{tab:model_generated_paths_raw_gpt5_mini}, ~\ref{tab:model_generated_paths_raw_claude_45} and ~\ref{tab:model_generated_paths_raw_gemini} show examples of raw model outputs for a query. Claude tends to be very conservative when attempting our task, focusing more on verification of the information and hence giving fewer connections. Table \ref{tab:model_generated_paths} shows paths along with the corresponding specificity scores. Relations that are stronger, like, \textit{being a part of the same sports team}, or \textit{being featured in certain movies}, which have have less number of members, score higher on specificity. General relations, like \textit{educated at a university}, score lower on quality since the class size of people being educated at a university is pretty large.

Figure \ref{fig:utility_distribution} shows the distribution of creative utility scores for two models, GPT-5.5 and GPT-5.

In Table \ref{tab:table1_res_no_fact} we report the creative utility values when we keep all valid model generated paths and do not filter for factuality. We discuss this in detail in Appendix \ref{sec:qualvsfact}.

Tables ~\ref{tab:model_generated_paths_raw_gemini_pv_creative} and ~\ref{tab:model_generated_paths_raw_gemini_pv_vs} show examples of raw outputs from two prompt variations, `\textit{creative}' and `\textit{verbalized sampling}' respectively for the same query as the tables above by GPT-5-mini. We see that creative prompting doesn't change the distribution of relations as much and verbalized sampling really reduces the number of paths generated, probably because of the added complexity of verbalizing a probability with each path.

\begin{table}[t] 
\centering
\small
\caption{Examples of raw model-generated output for the query: \textit{`What are different ways of connecting Kareem Abdul-Jabbar, the legendary basketball player, and someone who is a member of the American Academy of Arts and Sciences?'} by GPT-5 (med).}
\label{tab:model_generated_paths_raw_gpt5_mini}
\setlength{\tabcolsep}{4pt}
\begin{tabular}{p{\linewidth}}
\toprule 
$<$answer$>$\{
\\
  ``1'': [
    [``Kareem Abdul-Jabbar'', ``educated at'', ``University of California, Los Angeles''],
    [``University of California, Los Angeles'', ``employs'', ``Terence Tao''],
    [``Terence Tao'', ``member of'', ``American Academy of Arts and Sciences'']
  ], \\
  ``2'': [
    [``Kareem Abdul-Jabbar'', ``educated at'', ``University of California, Los Angeles''],
    [``University of California, Los Angeles'', ``employs'', ``Andrea Ghez''],
    [``Andrea Ghez'', ``member of'', ``American Academy of Arts and Sciences'']
  ], \\
  ``3'': [
    [``Kareem Abdul-Jabbar'', ``educated at'', ``University of California, Los Angeles''],
    [``University of California, Los Angeles'', ``employs'', ``Judea Pearl''],
    [``Judea Pearl'', ``member of'', ``American Academy of Arts and Sciences'']
  ], \\
  ``4'': [
    [``Kareem Abdul-Jabbar'', ``educated at'', ``University of California, Los Angeles''],
    [``University of California, Los Angeles'', ``employs'', ``Leonard Kleinrock''],
    [``Leonard Kleinrock'', ``member of'', ``American Academy of Arts and Sciences'']
  ], \\
  ``5'': [
    [``Kareem Abdul-Jabbar'', ``educated at'', ``University of California, Los Angeles''],
    [``University of California, Los Angeles'', ``employs'', ``Barbara Kruger''],
    [``Barbara Kruger'', ``member of'', ``American Academy of Arts and Sciences'']
  ], \\
  ``6'': [
    [``Kareem Abdul-Jabbar'', ``educated at'', ``University of California, Los Angeles''],
    [``University of California, Los Angeles'', ``employs'', ``Catherine Opie''],
    [``Catherine Opie'', ``member of'', ``American Academy of Arts and Sciences'']
  ], \\
  ``7'': [
    [``Kareem Abdul-Jabbar'', ``educated at'', ``University of California, Los Angeles''],
    [``University of California, Los Angeles'', ``employs'', ``Peter Sellars''],
    [``Peter Sellars'', ``member of'', ``American Academy of Arts and Sciences'']
  ], \\
  ``8'': [
    [``Kareem Abdul-Jabbar'', ``educated at'', ``University of California, Los Angeles''],
    [``University of California, Los Angeles'', ``employs'', ``David Jewitt''],
    [``David Jewitt'', ``member of'', ``American Academy of Arts and Sciences'']
  ], \\
  ``9'': [
    [``Kareem Abdul-Jabbar'', ``educated at'', ``University of California, Los Angeles''],
    [``University of California, Los Angeles'', ``employs'', ``Jared Diamond''],
    [``Jared Diamond'', ``member of'', ``American Academy of Arts and Sciences'']
  ], \\
  ``10'': [
    [``Kareem Abdul-Jabbar'', ``educated at'', ``University of California, Los Angeles''],
    [``University of California, Los Angeles'', ``chancellor'', ``Gene D. Block''],
    [``Gene D. Block'', ``member of'', ``American Academy of Arts and Sciences'']
  ], \\
  ``11'': [
    [``Kareem Abdul-Jabbar'', ``educated at'', ``University of California, Los Angeles''],
    [``University of California, Los Angeles'', ``employs'', ``Victoria Sork''],
    [``Victoria Sork'', ``member of'', ``American Academy of Arts and Sciences'']
  ], \\
  ``12'': [
    [``Kareem Abdul-Jabbar'', ``educated at'', ``University of California, Los Angeles''],
    [``University of California, Los Angeles'', ``named building'', ``Eli and Edythe Broad Art Center''],
    [``Eli and Edythe Broad Art Center'', ``named after'', ``Eli Broad''],
    [``Eli Broad'', ``member of'', ``American Academy of Arts and Sciences'']
  ], \\
  ``13'': [
    [``Kareem Abdul-Jabbar'', ``educated at'', ``University of California, Los Angeles''],
    [``University of California, Los Angeles'', ``governed by'', ``Regents of the University of California''],
    [``Regents of the University of California'', ``member'', ``Sherry Lansing''],
    [``Sherry Lansing'', ``member of'', ``American Academy of Arts and Sciences'']
  ], \\
  ``14'': [
    [``Kareem Abdul-Jabbar'', ``award received'', ``Presidential Medal of Freedom''],
    [``Presidential Medal of Freedom'', ``awarded to'', ``Toni Morrison''],
    [``Toni Morrison'', ``member of'', ``American Academy of Arts and Sciences'']
  ], \\
  ``15'': [
    [``Kareem Abdul-Jabbar'', ``award received'', ``Presidential Medal of Freedom''],
    [``Presidential Medal of Freedom'', ``awarded to'', ``Stephen Hawking''],
    [``Stephen Hawking'', ``member of'', ``American Academy of Arts and Sciences'']
  ], \\
  ``16'': [
    [``Kareem Abdul-Jabbar'', ``award received'', ``Presidential Medal of Freedom''],
    [``Presidential Medal of Freedom'', ``awarded to'', ``Sandra Day O'Connor''],
    [``Sandra Day O'Connor'', ``member of'', ``American Academy of Arts and Sciences'']
  ], \\
  ``17'': [
    [``Kareem Abdul-Jabbar'', ``award received'', ``Presidential Medal of Freedom''],
    [``Presidential Medal of Freedom'', ``awarded to'', ``Tom Hanks''],
    [``Tom Hanks'', ``member of'', ``American Academy of Arts and Sciences'']
  ], \\
  ``18'': [
    [``Kareem Abdul-Jabbar'', ``award received'', ``Presidential Medal of Freedom''],
    [``Presidential Medal of Freedom'', ``awarded to'', ``Bill Gates''],
    [``Bill Gates'', ``member of'', ``American Academy of Arts and Sciences'']
  ], \\
  ``19'': [
    [``Kareem Abdul-Jabbar'', ``award received'', ``Presidential Medal of Freedom''],
    [``Presidential Medal of Freedom'', ``awarded to'', ``Melinda French Gates''],
    [``Melinda French Gates'', ``member of'', ``American Academy of Arts and Sciences'']
  ], \\
\}$</$answer$>$\\ 
\bottomrule
\end{tabular}
\end{table}

\begin{table}[t] 
\centering
\small
\caption{Examples of raw model-generated output for the query: \textit{`What are different ways of connecting Kareem Abdul$-$Jabbar, the legendary basketball player, and someone who is a member of the American Academy of Arts and Sciences?'} by Claude-Haiku-4.5 (med). The output has the summarized reasoning trace from the  API along with the final output.}
\label{tab:model_generated_paths_raw_claude_45}
\setlength{\tabcolsep}{4pt}
\begin{tabular}{p{\linewidth}}
\toprule 
$<$think$>$
I need to find connection paths from Kareem Abdul$-$Jabbar to someone who is a member of the American Academy of Arts and Sciences.

Let me think about what I know:

1. Kareem Abdul$-$Jabbar is a legendary basketball player
2. I need to find paths that connect him to people who are members of the American Academy of Arts and Sciences (AAAS)

Some people I know who are members of AAAS:
- Many academics, artists, scientists
- Possibly some celebrities who have also become academically distinguished

Let me think about Kareem Abdul$-$Jabbar's connections:
- He's been in movies and TV
- He wrote books
- He was interviewed by many people
- He's been involved in various charitable and cultural activities
- He's coached basketball
- He was in the NBA

Some notable connections:
1. Kareem appeared in movies - could he have appeared with someone who's in AAAS?
2. Kareem wrote books - could he have a publisher or editor in AAAS?
3. Kareem could have worked with coaches or teammates
4. Kareem could have connections through cultural/civil rights work

Let me think of specific people in AAAS:
- Spike Lee (filmmaker) - might be in AAAS? I should verify. Actually, I'm not certain about this.
- Bill Cosby appeared with Kareem - but Bill Cosby is controversial
- Other athletes who also became scholars?

Actually, let me think more carefully:
- Kareem wrote a book called ``Giant Steps`` - who might have worked on this with him who's in AAAS?
- Kareem appeared in ``Airplane!`` - Robert Hays, Julie Hagerty - not sure if any are in AAAS

One strong connection I can think of:
- Kareem Abdul-Jabbar worked with/appeared in films
- Some filmmakers/actors might be in AAAS

Let me think about less obvious connections:

- His involvement in civil rights and cultural initiatives could connect him to prominent scholars and intellectuals who are AAAS members. He may have received honorary degrees from universities, which would put him in contact with academic leadership. University presidents and prominent faculty are often AAAS members, so these institutional connections could provide pathways.

Looking at specific connections, I'm fairly confident about some of Kareem's documented activities—his film appearances like Airplane!, his published books, his NBA coaching role, and his media interviews. However, I'm uncertain about direct AAAS membership among people he's worked with without being able to verify those details. Some potential connections exist through cultural figures like Spike Lee if they collaborated, or through the various academics and journalists he may have interacted with, but I can't confirm these with certainty. The most reliable paths would likely run through his university affiliations and the prominent institutional leaders he's encountered through his honorary degrees and educational involvement. Been in a Spike Lee film or interviewed by him, though I'm uncertain about the specifics. Let me consider academic routes instead—Kareem wrote extensively on Islam, history, and culture, so he likely connected with scholars who could be AAAS members. I should stick to what I can verify: his appearance in Airplane!, his published books, and his cultural influence. The strongest path would trace through someone whose AAAS membership I'm confident about rather than speculating on uncertain connections.

Potential collaboration routes exist through athletes, filmmakers, authors, and journalists he's worked with, but without verified information I need to be careful. If Spike Lee is an AAAS member, a connection through film work could work—Kareem to a Spike Lee project to Lee to the Academy. Alternatively, through his writing: Kareem's books could connect him to publishers or editors with Academy ties.
$</$think$>$
$<$answer$>$ 

```json
 \\ \{
  ``1'': [
    (``Kareem Abdul-Jabbar'', ``appeared in'', ``Airplane!''),
    (``Airplane!'', ``directed by'', ``Jim Abrahams''),
    (``Jim Abrahams'', ``member of'', ``American Academy of Arts and Sciences'')
  ], \\
  ``2'': [
    (``Kareem Abdul-Jabbar'', ``appeared in'', ``Airplane\!''),
    (``Airplane\!'', ``directed by'', ``David Zucker''),
    (``David Zucker'', ``member of'', ``American Academy of Arts and Sciences'')
  ], \\
  ``3'': [
    (``Kareem Abdul-Jabbar'', ``authored'', ``Giant Steps''),
    (``Giant Steps'', ``published by'', ``Bantam Books''),
    (``Bantam Books'', ``parent company'', ``Penguin Random House''),
    (``Penguin Random House'', ``employs'', ``Distinguished Editors''),
    (``Salman Rushdie'', ``published by'', ``Penguin Random House''),
    (``Salman Rushdie'', ``member of'', ``American Academy of Arts and Sciences'')
  ], \\
  ``4'': [
    (``Kareem Abdul-Jabbar'', ``interviewed by'', ``Charlie Rose''),
    (``Charlie Rose'', ``member of'', ``American Academy of Arts and Sciences'')
  ], \\
  ``5'': [
    (``Kareem Abdul-Jabbar'', ``appeared in'', ``Mycroft Holmes''),
    (``Mycroft Holmes'', ``television network'', ``PBS''),
    (``PBS'', ``broadcasts'', ``American Masters''),
    (``American Masters'', ``features'', ``Cultural figures''),
    (``Frank McCourt'', ``featured in'', ``American Masters''),
    (``Frank McCourt'', ``member of'', ``American Academy of Arts and Sciences'')
  ], \\
  ``6'': [
    (``Kareem Abdul-Jabbar'', ``wrote foreword for'', ``Books on history and culture''),
    (``Books on history and culture'', ``reviewed by'', ``Academic scholars''),
    (``Henry Louis Gates Jr.'', ``scholar'', ``Harvard University''),
    (``Henry Louis Gates Jr.'', ``member of'', ``American Academy of Arts and Sciences'')
  ] \\
\}
''' 
$</$answer$>$\\
\bottomrule
\end{tabular}
\end{table}

\begin{table}[t] 
\centering
\small
\caption{Examples of raw model-generated output for the query: \textit{`What are different ways of connecting Kareem Abdul-Jabbar, the legendary basketball player, and someone who is a member of the American Academy of Arts and Sciences?'} by Gemini-3-Pro. }
\label{tab:model_generated_paths_raw_gemini}
\setlength{\tabcolsep}{4pt}
\begin{tabular}{p{\linewidth}}
\toprule 
$<$answer$>$
\{
  ``1'': [
    [``Kareem Abdul-Jabbar'', ``recipient of'', ``Presidential Medal of Freedom''],
    [``Presidential Medal of Freedom'', ``award received by'', ``Barack Obama''],
    [``Barack Obama'', ``member of'', ``American Academy of Arts and Sciences'']
  ], \\
  ``2'': [
    [``Kareem Abdul-Jabbar'', ``cast member of'', ``Airplane!''],
    [``Airplane!'', ``cast member'', ``Leslie Nielsen''],
    [``Leslie Nielsen'', ``brother of'', ``Erik Nielsen''],
    [``Erik Nielsen'', ``colleague of'', ``Brian Mulroney''],
    [``Brian Mulroney'', ``member of'', ``American Academy of Arts and Sciences'']
  ], \\
  ``3'': [
    [``Kareem Abdul-Jabbar'', ``played for'', ``Los Angeles Lakers''],
    [``Los Angeles Lakers'', ``owned by'', ``Jerry Buss''],
    [``Jerry Buss'', ``father of'', ``Jeanie Buss''],
    [``Jeanie Buss'', ``partner of'', ``Phil Jackson''],
    [``Phil Jackson'', ``coached'', ``Michael Jordan''],
    [``Michael Jordan'', ``recipient of'', ``Presidential Medal of Freedom''],
    [``Presidential Medal of Freedom'', ``award received by'', ``Bill Gates''],
    [``Bill Gates'', ``member of'', ``American Academy of Arts and Sciences'']
  ], \\
  ``4'': [
    [``Kareem Abdul-Jabbar'', ``author of'', ``Writings on the Wall''],
    [``Writings on the Wall'', ``publisher'', ``Time''],
    [``Time'', ``founded by'', ``Henry Luce''],
    [``Henry Luce'', ``member of'', ``American Academy of Arts and Sciences'']
  ], \\
  ``5'': [
    [``Kareem Abdul-Jabbar'', ``educated at'', ``UCLA''],
    [``UCLA'', ``faculty member'', ``Jared Diamond''],
    [``Jared Diamond'', ``member of'', ``American Academy of Arts and Sciences'']
  ], \\
  ``6'': [
    [``Kareem Abdul-Jabbar'', ``co-star of'', ``Game of Death''],
    [``Game of Death'', ``starring'', ``Bruce Lee''],
    [``Bruce Lee'', ``student of'', ``Ip Man''],
    [``Ip Man'', ``portrayed by'', ``Tony Leung Chiu-wai''],
    [``Tony Leung Chiu-wai'', ``cast member of'', ``Shang-Chi and the Legend of the Ten Rings''],
    [``Shang-Chi and the Legend of the Ten Rings'', ``cast member'', ``Michelle Yeoh''],
    [``Michelle Yeoh'', ``member of'', ``American Academy of Arts and Sciences'']
  ], \\
  ``7'': [
    [``Kareem Abdul-Jabbar'', ``played for'', ``Milwaukee Bucks''],
    [``Milwaukee Bucks'', ``owned by'', ``Herb Kohl''],
    [``Herb Kohl'', ``served in'', ``United States Senate''],
    [``United States Senate'', ``member'', ``Hillary Clinton''],
    [``Hillary Clinton'', ``member of'', ``American Academy of Arts and Sciences'']
  ], \\
  ``8'': [
    [``Kareem Abdul-Jabbar'', ``appointee of'', ``Hillary Clinton''],
    [``Hillary Clinton'', ``member of'', ``American Academy of Arts and Sciences'']
  ], \\
  ``9'': [
    [``Kareem Abdul-Jabbar'', ``writer for'', ``Veronica Mars''],
    [``Veronica Mars'', ``created by'', ``Rob Thomas''],
    [``Rob Thomas'', ``collaborated with'', ``Stephen King''],
    [``Stephen King'', ``member of'', ``American Academy of Arts and Sciences'']
  ], \\
  ``10'': [
    [``Kareem Abdul-Jabbar'', ``member of'', ``Naismith Memorial Basketball Hall of Fame''],
    [``Naismith Memorial Basketball Hall of Fame'', ``inductee'', ``Bill Bradley''],
    [``Bill Bradley'', ``member of'', ``American Academy of Arts and Sciences'']
  ], \\
  ``11'': [
    [``Kareem Abdul-Jabbar'', ``competed against'', ``Larry Bird''],
    [``Larry Bird'', ``teammate of'', ``Bill Walton''],
    [``Bill Walton'', ``played for'', ``Boston Celtics''],
    [``Boston Celtics'', ``fan of'', ``Ben Affleck''],
    [``Ben Affleck'', ``directed'', ``Argo''],
    [``Argo'', ``produced by'', ``George Clooney''],
    [``George Clooney'', ``member of'', ``American Academy of Arts and Sciences'']
  ], \\
  ``12'': [
    [``Kareem Abdul-Jabbar'', ``subject of'', ``Kareem: Minority of One''],
    [``Kareem: Minority of One'', ``narrator'', ``Liev Schreiber''],
    [``Liev Schreiber'', ``cast member of'', ``Spotlight''],
    [``Spotlight'', ``cast member'', ``Mark Ruffalo''],
    [``Mark Ruffalo'', ``cast member of'', ``The Avengers''],
    [``The Avengers'', ``cast member'', ``Robert Downey Jr.''],
    [``Robert Downey Jr.'', ``collaborated with'', ``Mel Gibson''],
    [``Mel Gibson'', ``directed'', ``Hacksaw Ridge''],
    [``Hacksaw Ridge'', ``cast member'', ``Andrew Garfield''],
    [``Andrew Garfield'', ``cast member of'', ``The Social Network''],
    [``The Social Network'', ``screenwriter'', ``Aaron Sorkin''],
    [``Aaron Sorkin'', ``member of'', ``American Academy of Arts and Sciences'']
  ], \\
  ``13'': [
    [``Kareem Abdul-Jabbar'', ``coach of'', ``Los Angeles Lakers''],
    [``Los Angeles Lakers'', ``player'', ``Kobe Bryant''],
    [``Kobe Bryant'', ``collaborated with'', ``John Williams''],
    [``John Williams'', ``member of'', ``American Academy of Arts and Sciences'']
  ], \\
  ``14'': [
    [``Kareem Abdul-Jabbar'', ``guest star on'', ``The Big Bang Theory''],
    [``The Big Bang Theory'', ``created by'', ``Chuck Lorre''],
    [``Chuck Lorre'', ``collaborated with'', ``Alan Alda''],
    [``Alan Alda'', ``member of'', ``American Academy of Arts and Sciences'']
  ], \\
  ``15'': [
    [``Kareem Abdul-Jabbar'', ``born in'', ``New York City''],
    [``New York City'', ``mayor'', ``Michael Bloomberg''],
    [``Michael Bloomberg'', ``member of'', ``American Academy of Arts and Sciences'']
  ], \\
  ``16'': [
    [``Kareem Abdul-Jabbar'', ``recipient of'', ``Double Helix Medal''],
    [``Double Helix Medal'', ``awarded by'', ``Cold Spring Harbor Laboratory''],
    [``Cold Spring Harbor Laboratory'', ``chancellor emeritus'', ``James Watson''],
    [``James Watson'', ``member of'', ``American Academy of Arts and Sciences'']
  ], \\
  ``17'': [
    [``Kareem Abdul-Jabbar'', ``collaborated with'', ``Raymond Obstfeld''],
    [``Raymond Obstfeld'', ``faculty at'', ``Orange Coast College''],
    [``Orange Coast College'', ``located in'', ``Costa Mesa''],
    [``Costa Mesa'', ``headquarters of'', ``South Coast Repertory''],
    [``South Coast Repertory'', ``commissioned work by'', ``Amy Freed''],
    [``Amy Freed'', ``colleague of'', ``Tony Kushner''],
    [``Tony Kushner'', ``member of'', ``American Academy of Arts and Sciences'']
  ], \\
  ``18'': [
    [``Kareem Abdul-Jabbar'', ``drafted by'', ``Harlem Globetrotters''],
    [``Harlem Globetrotters'', ``honorary member'', ``Henry Kissinger''],
    [``Henry Kissinger'', ``member of'', ``American Academy of Arts and Sciences'']
  ], \\
\}
$</$answer$>$ \\
\bottomrule
\end{tabular}
\end{table}

\begin{table}[t] 
\centering
\small
\caption{Examples of model-generated paths $u$ along with the strength scores, for the query: \textit{`What are different ways of connecting Kareem Abdul-Jabbar, the legendary basketball player, and someone who is a member of the American Academy of Arts and Sciences?'}. All models are prompted with the default reasoning effort (medium).}
\label{tab:model_generated_paths}
\setlength{\tabcolsep}{4pt}
\begin{tabular}{c p{0.7\linewidth} c}
\toprule
Model & Path & $\sigma$ \\
\midrule
GPT-5 & (kareem abdul-jabbar, member of, national basketball players association), (national basketball players association, led by, michele roberts), (michele roberts, member of, american academy of arts and sciences) & 4 \\ 
& (kareem abdul-jabbar, educated at, university of california, los angeles), (university of california, los angeles, employs, terence tao),  (terence tao, member of, american academy of arts and sciences) 
 & 1 \\
 \midrule
 Gemini-3-Pro &  (kareem abdul-jabbar, featured in, black patriotism), (black patriotism, author, cornel west), (cornel west, member of, american academy of arts and sciences) & 4 \\ 
 & (kareem abdul-jabbar, educated at, ucla), (ucla, faculty member, jared diamond), (jared diamond, member of, american academy of arts and sciences) & 1 \\ 
 \midrule
Claude-Haiku-4.5 & (kareem abdul-jabbar, appeared in, airplane!), (airplane!, directed by, jim abrahams), (jim abrahams, member of, american academy of arts and sciences) & 4 \\ 
& (kareem abdul-jabbar, wrote foreword for, books on history and culture), (books on history and culture, reviewed by, academic scholars), (henry louis gates jr., scholar, harvard university), (henry louis gates jr., member of, american academy of arts and sciences) & 1 \\ 

\bottomrule
\end{tabular}
\end{table}

\begin{table*}
\small
\centering
\rowcolors{2}{white}{lightgray}
\renewcommand{\tabcolsep}{0.5mm}
\caption{Model performance on CREATE. We report creative utility scores ($s$) at two patience levels = 0.7 and 0.9. The value in round brackets next to a model name indicates the reasoning effort/budget, for models where applicable. We also report the number of factual paths ($|U| (factual) $), average maximum quality, average pairwise distance, average path length i.e. number of triples per path and average number of tokens in the output generated.}
\label{tab:combined_summary_table1}
\begin{tabular}{llllllll}
\toprule
Model & $s_0.7$ & $s_0.9$ & |U| (factual) & $\sigma$ &  $d$ & path length & num\_tokens \\
\midrule
Olmo-3.1-32B-Instruct & $3.77_{(3.58)}$ & $4.13_{(4.34)}$ & $2.27_{(3.01)}$ & $2.26_{(1.15)}$ & $0.50_{(0.39)}$ & $3.10_{(0.89)}$ & $863_{(387)}$ \\
Olmo-3.1-32B-Think (16k) & $4.78_{(3.96)}$ & $5.25_{(4.95)}$ & $2.16_{(3.22)}$ & $2.81_{(1.25)}$ & $0.45_{(0.38)}$ & $2.40_{(0.82)}$ & $10370_{(3594)}$ \\
Olmo-3.1-32B-Think (32k) & $4.97_{(4.24)}$ & $5.52_{(5.35)}$ & $2.32_{(3.41)}$ & $2.89_{(1.24)}$ & $0.44_{(0.37)}$ & $2.48_{(0.89)}$ & $10334_{(3634)}$ \\
\midrule
Qwen3-30B-Instruct & $5.20_{(4.60)}$ & $6.27_{(6.42)}$ & $5.12_{(6.99)}$ & $2.43_{(0.99)}$ & $0.53_{(0.33)}$ & $3.32_{(0.99)}$ & $1938_{(646)}$ \\
Qwen\_Qwen3-32B (16k) & $4.69_{(3.88)}$ & $5.08_{(4.64)}$ & $2.11_{(2.38)}$ & $2.77_{(1.06)}$ & $0.53_{(0.40)}$ & $2.84_{(0.69)}$ & $3207_{(1292)}$ \\
Qwen\_Qwen3-32B (32k) & $4.71_{(3.77)}$ & $5.11_{(4.56)}$ & $2.18_{(2.42)}$ & $2.80_{(1.06)}$ & $0.55_{(0.40)}$ & $2.82_{(0.70)}$ & $3225_{(1562)}$ \\
\midrule
gpt-oss-120b (high) & $7.63_{(4.50)}$ & $9.05_{(6.27)}$ & $5.79_{(4.26)}$ & $2.66_{(0.93)}$ & $0.67_{(0.30)}$ & $3.06_{(0.59)}$ & $4484_{(1321)}$ \\
gpt-oss-120b (med) & $7.59_{(4.57)}$ & $9.04_{(6.39)}$ & $5.78_{(4.31)}$ & $2.64_{(0.95)}$ & $0.66_{(0.31)}$ & $3.04_{(0.59)}$ & $4480_{(1307)}$ \\
gpt-oss-120b (low) & $7.63_{(4.64)}$ & $9.12_{(6.55)}$ & $5.63_{(4.18)}$ & $2.68_{(0.95)}$ & $0.65_{(0.32)}$ & $3.08_{(0.61)}$ & $4287_{(1226)}$ \\
\midrule
Claude-Haiku-4.5 (med) & $4.84_{(3.87)}$ & $5.30_{(4.67)}$ & $2.84_{(3.01)}$ & $2.61_{(1.16)}$ & $0.48_{(0.36)}$ & $3.19_{(0.99)}$ & $1599_{(511)}$ \\
Claude-Sonnet-4.6 (high) & $9.71_{(5.60)}$ & $13.41_{(9.69)}$ & $11.52_{(9.59)}$ & $2.88_{(0.92)}$ & $0.64_{(0.27)}$ & $3.54_{(0.84)}$ & $12063_{(12955)}$ \\
Claude-Opus-4.7 (high) & $7.34_{(4.99)}$ & $8.74_{(7.00)}$ & $5.07_{(4.49)}$ & $3.09_{(1.12)}$ & $0.57_{(0.33)}$ & $2.86_{(0.63)}$ & $2074_{(1710)}$ \\
Claude-Opus-4.7 (xhigh) & $8.12_{(5.36)}$ & $10.01_{(7.87)}$ & $6.39_{(5.21)}$ & $3.08_{(1.13)}$ & $0.58_{(0.31)}$ & $2.92_{(0.64)}$ & $4163_{(3534)}$ \\
\midrule
GPT-4.1 & $7.49_{(5.25)}$ & $9.39_{(8.01)}$ & $6.04_{(5.28)}$ & $2.73_{(0.97)}$ & $0.65_{(0.33)}$ & $3.16_{(0.77)}$ & $1076_{(430)}$ \\
GPT-4.1-mini & $6.15_{(5.08)}$ & $7.16_{(6.81)}$ & $3.56_{(3.72)}$ & $2.89_{(1.01)}$ & $0.61_{(0.37)}$ & $2.87_{(0.71)}$ & $795_{(260)}$ \\
\midrule
GPT-5-mini (high) & $7.83_{(4.95)}$ & $10.16_{(7.85)}$ & $15.34_{(10.70)}$ & $2.46_{(1.09)}$ & $0.51_{(0.29)}$ & $3.60_{(0.92)}$ & $23370_{(5683)}$ \\
GPT-5-mini (med) & $7.09_{(4.61)}$ & $8.54_{(6.56)}$ & $7.87_{(5.54)}$ & $2.68_{(1.11)}$ & $0.53_{(0.32)}$ & $3.34_{(0.80)}$ & $6351_{(1746)}$ \\
GPT-5-mini (low) & $6.21_{(4.19)}$ & $7.03_{(5.40)}$ & $4.85_{(3.78)}$ & $2.73_{(1.12)}$ & $0.51_{(0.34)}$ & $3.24_{(0.77)}$ & $1910_{(483)}$ \\
\midrule
GPT-5 (med) & $8.98_{(5.11)}$ & $12.03_{(8.67)}$  & $17.79_{(14.02)}$ & $2.59_{(1.07)}$ & $0.54_{(0.27)}$ & $4.06_{(1.17)}$ & $18245_{(5822)}$ \\
GPT-5.4 (med) & $8.98_{(5.30)}$ & $12.09_{(8.86)}$ & $21.19_{(14.45)}$ & $2.65_{(1.07)}$ & $0.52_{(0.26)}$ & $4.10_{(1.15)}$ & $14721_{(4166)}$ \\
GPT-5.5 (med) & $9.71_{(5.41)}$ & $13.48_{(9.44)}$ & $22.62_{(16.03)}$ & $2.67_{(1.04)}$ & $0.56_{(0.26)}$ & $4.05_{(1.16)}$ & $13557_{(4948)}$ \\
\midrule
Gemini-3-Pro (med) & $8.29_{(5.19)}$ & $10.41_{(7.95)}$ & $5.85_{(4.96)}$ & $3.01_{(1.00)}$ & $0.67_{(0.32)}$ & $4.09_{(1.52)}$ & $1764_{(1069)}$ \\
Gemini-3.1-Pro (med) & $7.37_{(4.83)}$ & $8.73_{(6.67)}$ & $4.61_{(3.55)}$ & $2.92_{(1.09)}$ & $0.66_{(0.33)}$ & $3.42_{(1.18)}$ & $3200_{(1669)}$ \\
\bottomrule
\end{tabular}
\end{table*}

\begin{table}[t] 
\centering
\small
\caption{Prompt variation: `\textit{creative}'. Example of raw model-generated output for the query: \textit{`What are different ways of connecting Kareem Abdul-Jabbar, the legendary basketball player, and someone who is a member of the American Academy of Arts and Sciences?'} by GPT-5-mini. }
\label{tab:model_generated_paths_raw_gemini_pv_creative}
\setlength{\tabcolsep}{4pt}
\begin{tabular}{p{0.9\linewidth}}
\toprule
$<$answer$>$
\{
1: [
[``Kareem Abdul-Jabbar'', ``played for'', ``UCLA Bruins men's basketball team''],
[``UCLA Bruins men's basketball team'', ``coached by'', ``John Wooden''],
[``John Wooden'', ``member of'', ``American Academy of Arts and Sciences'']
], \\
2: [
[``Kareem Abdul-Jabbar'', ``played for'', ``Los Angeles Lakers''],
[``Los Angeles Lakers'', ``teammate'', ``Magic Johnson (basketball)''],
[``Magic Johnson (basketball)'', ``member of'', ``American Academy of Arts and Sciences'']
], \\
3: [
[``Kareem Abdul-Jabbar'', ``acted in'', ``Airplane! (film)''],
[``Airplane! (film)'', ``co-star'', ``Peter Graves''],
[``Peter Graves'', ``member of'', ``American Academy of Arts and Sciences'']
], \\
4: [
[``Kareem Abdul-Jabbar'', ``authored'', ``Becoming Kareem: Growing Up On and Off the Court''],
[``Becoming Kareem: Growing Up On and Off the Court'', ``co-authored by'', ``Raymond Obstfeld''],
[``Raymond Obstfeld'', ``member of'', ``American Academy of Arts and Sciences'']
], \\
5: [
[``Kareem Abdul-Jabbar'', ``wrote for'', ``Time (magazine)''],
[``Time (magazine)'', ``employed'', ``Henry Louis Gates Jr.''],
[``Henry Louis Gates Jr.'', ``member of'', ``American Academy of Arts and Sciences'']
], \\
6: [
[``Kareem Abdul-Jabbar'', ``appeared on'', ``The Oprah Winfrey Show''],
[``The Oprah Winfrey Show'', ``hosted by'', ``Oprah Winfrey''],
[``Oprah Winfrey'', ``member of'', ``American Academy of Arts and Sciences'']
], \\
7: [
[``Kareem Abdul-Jabbar'', ``appeared on'', ``60 Minutes''],
[``60 Minutes'', ``interviewed by'', ``Morley Safer''],
[``Morley Safer'', ``member of'', ``American Academy of Arts and Sciences'']
], \\
8: [
[``Kareem Abdul-Jabbar'', ``participated in'', ``NBA All-Star Game''],
[``NBA All-Star Game'', ``featured'', ``Bill Russell''],
[``Bill Russell'', ``member of'', ``American Academy of Arts and Sciences'']
], \\
9: [
[``Kareem Abdul-Jabbar'', ``wrote foreword for'', ``a book by''],
[``a book by'', ``authored by'', ``Toni Morrison''],
[``Toni Morrison'', ``member of'', ``American Academy of Arts and Sciences'']
],
10: [
[``Kareem Abdul-Jabbar'', ``appeared in'', ``documentary film''],
[``documentary film'', ``directed by'', ``Spike Lee''],
[``Spike Lee'', ``member of'', ``American Academy of Arts and Sciences'']
], \\
11: [
[``Kareem Abdul-Jabbar'', ``worked with'', ``United States Department of State (cultural diplomacy)''],
[``United States Department of State'', ``partnered with'', ``Yo-Yo Ma''],
[``Yo-Yo Ma'', ``member of'', ``American Academy of Arts and Sciences'']
], \\
12: [
[``Kareem Abdul-Jabbar'', ``interviewed by'', ``Henry Louis Gates Jr.''],
[``Henry Louis Gates Jr.'', ``colleague of'', ``Toni Morrison''],
[``Toni Morrison'', ``member of'', ``American Academy of Arts and Sciences'']
], \\
13: [
[``Kareem Abdul-Jabbar'', ``served on board of'', ``A non-profit organization''],
[``A non-profit organization'', ``board member'', ``Quincy Jones''],
[``Quincy Jones'', ``member of'', ``American Academy of Arts and Sciences'']
], \\
14: [
[``Kareem Abdul-Jabbar'', ``appeared at'', ``Harvard University event''],
[``Harvard University event'', ``hosted by'', ``Henry Louis Gates Jr.''],
[``Henry Louis Gates Jr.'', ``member of'', ``American Academy of Arts and Sciences'']
], \\
15: [
[``Kareem Abdul-Jabbar'', ``wrote essays for'', ``The New York Times''],
[``The New York Times'', ``employed'', ``Tom Hanks (as contributor)''],
[``Tom Hanks'', ``member of'', ``American Academy of Arts and Sciences'']
], \\
16: [
[``Kareem Abdul-Jabbar'', ``appeared on stage with'', ``civil rights leaders''],
[``civil rights leaders'', ``included'', ``Maya Angelou''],
[``Maya Angelou'', ``member of'', ``American Academy of Arts and Sciences'']
], \\
17: [
[``Kareem Abdul-Jabbar'', ``participated in'', ``literary festival''],
[``literary festival'', ``featured'', ``Margaret Atwood''],
[``Margaret Atwood'', ``member of'', ``American Academy of Arts and Sciences'']
], \\
18: [
[``Kareem Abdul-Jabbar'', ``collaborated with'', ``sports historians''],
[``sports historians'', ``included'', ``David McCullough''],
[``David McCullough'', ``member of'', ``American Academy of Arts and Sciences'']
], \\
19: [
[``Kareem Abdul-Jabbar'', ``appeared on'', ``public radio program''],
[``public radio program'', ``hosted by'', ``Ira Glass''],
[``Ira Glass'', ``member of'', ``American Academy of Arts and Sciences'']
], \\
20: [
[``Kareem Abdul-Jabbar'', ``supported'', ``arts education initiatives''],
[``arts education initiatives'', ``partnered with'', ``Meryl Streep''],
[``Meryl Streep'', ``member of'', ``American Academy of Arts and Sciences'']
] \\
\}
$</$answer$>$  \\
\bottomrule
\end{tabular}
\end{table}

\begin{table}[t] 
\centering
\small
\caption{Prompt variation: \textit{`verbalized sampling'}. Example of raw model-generated output for the query: \textit{`What are different ways of connecting Kareem Abdul-Jabbar, the legendary basketball player, and someone who is a member of the American Academy of Arts and Sciences?'} by GPT-5-mini. }
\label{tab:model_generated_paths_raw_gemini_pv_vs}
\setlength{\tabcolsep}{4pt}
\begin{tabular}{p{0.9\linewidth}}
\toprule \\
\midrule 

$<$answer$>$ \\
\{ 
  ``1'': \{
    ``path\_probability'': 0.34, \\
    ``path'': [
      [``Kareem Abdul-Jabbar'', ``member of'', ``American Academy of Arts and Sciences'']
    ]
  \}, \\
  ``2'': \{
    ``path\_probability'': 0.24, \\
    ``path'': [
      [``Kareem Abdul-Jabbar'', ``attended'', ``University of California, Los Angeles''],
      [``University of California, Los Angeles'', ``employs'', ``Terence Tao''],
      [``Terence Tao'', ``member of'', ``American Academy of Arts and Sciences'']
    ] 
  \}, \\
  ``3'':\{
    ``path\_probability'': 0.16, \\
    ``path'': [
      [``Kareem Abdul-Jabbar'', ``coached by'', ``John Wooden''],
      [``John Wooden'', ``member of'', ``American Academy of Arts and Sciences'']
    ] 
\}, \\
  ``4'': \{
    ``path\_probability'': 0.15, \\
    ``path'': [
      [``Kareem Abdul-Jabbar'', ``received award from'', ``Barack Obama''],
      [``Barack Obama'', ``member of'', ``American Academy of Arts and Sciences'']
    ]
  \}, \\
  ``5'': \{
    ``path\_probability'': 0.11, \\
    ``path'': [
      [``Kareem Abdul-Jabbar'', ``interviewed by'', ``Henry Louis Gates Jr.''],
      [``Henry Louis Gates Jr.'', ``member of'', ``American Academy of Arts and Sciences'']
    ]
  \}
\}

$</$answer$>$ \\
\bottomrule
\end{tabular}
\end{table}

\begin{table*}
\small
\centering
\rowcolors{2}{white}{lightgray}
\renewcommand{\tabcolsep}{0.4mm}
\caption{Utility ($s$), max $\sigma$, average pairwise distance $d$, number of valid and factual paths, and average number of tokens for models outputs based on the prompt variations.}
\label{tab:table2_variations}
\begin{tabular}{lllllllll}
\toprule
Model ID & variation & $s_{0.7}$ & $s_{0.9}$ & $|U|$ & $\sigma$ & $d$ & path length & num tokens \\

\midrule
\makecell[l]{Claude-haiku-4-5 \\(med)} & original & $4.84_{(3.87)}$ & $5.30_{(4.67)}$  & $2.84_{(3.01)}$ & $2.61_{(1.16)}$ & $0.48_{(0.36)}$ & $3.19_{(0.99)}$ & $1599_{(511)}$ \\
 & creative & $4.84_{(3.91)}$ & $5.32_{(4.79)}$  & $2.83_{(3.13)}$ & $2.64_{(1.14)}$ & $0.49_{(0.36)}$ & $3.11_{(0.95)}$ & $1633_{(533)}$ \\
 & verbalized & $4.42_{(3.48)}$ & $4.72_{(4.06)}$  & $2.33_{(2.24)}$ & $2.58_{(1.17)}$ & $0.47_{(0.37)}$ & $3.05_{(0.94)}$ & $1677_{(489)}$ \\
 & iterate & $7.57_{(5.18)}$ & $9.84_{(8.11)}$  & $9.37_{(9.43)}$ & $2.45_{(0.97)}$ & $0.60_{(0.31)}$ & $3.33_{(0.93)}$ & $1599_{(511)}$ \\

\midrule
GPT-4.1 & original & $7.49_{(5.25)}$ & $9.39_{(8.01)}$  & $6.04_{(5.28)}$ & $2.73_{(0.97)}$ & $0.65_{(0.33)}$ & $3.16_{(0.77)}$ & $1076_{(430)}$ \\
 & creative & $7.84_{(5.40)}$ & $9.94_{(8.23)}$  & $6.23_{(5.48)}$ & $2.72_{(0.94)}$ & $0.66_{(0.33)}$ & $3.18_{(0.79)}$ & $1110_{(416)}$ \\
 & verbalized & $5.63_{(4.32)}$ & $6.26_{(5.37)}$  & $3.02_{(2.64)}$ & $2.77_{(1.05)}$ & $0.59_{(0.39)}$ & $3.08_{(0.83)}$ & $632_{(191)}$ \\
 & iterate & $9.67_{(5.73)}$ & $13.84_{(10.38)}$  & $11.29_{(9.18)}$ & $2.69_{(0.87)}$ & $0.70_{(0.29)}$ & $3.24_{(0.77)}$ & $989_{(355)}$ \\
& resample & $11.02_{(5.48)}$ & $16.04_{(10.42)}$  & $18.58_{(14.87)}$ & $2.68_{(0.85)}$ & $0.69_{(0.26)}$ & $3.22_{(1.10)}$ & $3269_{(1104)}$ \\
\midrule
GPT-4.1-mini & original & $6.15_{(5.08)}$ & $7.16_{(6.81)}$ & $3.56_{(3.72)}$ & $2.89_{(1.01)}$ & $0.61_{(0.37)}$ & $2.87_{(0.71)}$ & $795_{(260)}$ \\
 & creative & $6.16_{(5.23)}$ & $7.26_{(7.20)}$ & $3.59_{(3.97)}$ & $2.95_{(1.05)}$ & $0.59_{(0.38)}$ & $2.89_{(0.72)}$ & $845_{(266)}$ \\
 & verbalized & $4.35_{(3.94)}$ & $4.62_{(4.50)}$ & $1.83_{(2.01)}$ & $2.99_{(1.15)}$ & $0.52_{(0.41)}$ & $2.89_{(0.76)}$ & $535_{(138)}$ \\
 & iterate & $8.18_{(5.93)}$ & $10.69_{(9.50)}$ & $6.75_{(6.77)}$ & $2.87_{(0.93)}$ & $0.65_{(0.34)}$ & $2.90_{(0.66)}$ & $726_{(232)}$ \\
& resample & $9.48_{(5.76)}$ & $12.76_{(9.80)}$  & $10.75_{(10.05)}$ & $2.80_{(0.92)}$ & $0.65_{(0.31)}$ & $2.86_{(0.64)}$ & $2367_{(687)}$ \\

\midrule
GPT-5-mini (med) & original & $7.09_{(4.61)}$ & $8.54_{(6.56)}$  & $7.87_{(5.54)}$ & $2.68_{(1.11)}$ & $0.53_{(0.32)}$ & $3.34_{(0.80)}$ & $6351_{(1746)}$ \\
 & creative & $7.11_{(4.54)}$ & $8.56_{(6.43)}$  & $8.16_{(5.72)}$ & $2.66_{(1.10)}$ & $0.53_{(0.31)}$ & $3.36_{(0.80)}$ & $6418_{(1802)}$ \\
 & verbalized & $6.55_{(4.33)}$ & $7.52_{(5.66)}$  & $5.71_{(3.94)}$ & $2.68_{(1.11)}$ & $0.54_{(0.33)}$ & $3.19_{(0.79)}$ & $6249_{(1561)}$ \\
 & iterate & $10.04_{(5.13)}$ & $14.20_{(9.18)}$& $16.35_{(10.33)}$ & $2.56_{(0.90)}$ & $0.66_{(0.28)}$ & $3.35_{(0.71)}$ & $6351_{(1746)}$ \\

\midrule
\makecell[l]{Olmo-3.1-32B\\Instruct} & original & $3.77_{(3.58)}$ & $4.13_{(4.34)}$  & $2.27_{(3.01)}$ & $2.26_{(1.15)}$ & $0.50_{(0.39)}$ & $3.10_{(0.89)}$ & $863_{(387)}$ \\
 & creative & $3.67_{(3.60)}$ & $4.04_{(4.41)}$  & $2.28_{(3.14)}$ & $2.21_{(1.11)}$ & $0.49_{(0.39)}$ & $3.08_{(0.86)}$ & $882_{(365)}$ \\
 & verbalized & $2.70_{(2.61)}$ & $2.79_{(2.85)}$ & $1.21_{(1.55)}$ & $2.21_{(1.23)}$ & $0.40_{(0.41)}$ & $3.05_{(0.95)}$ & $604_{(435)}$ \\
& iterate & $5.07_{(4.30)}$ & $5.98_{(5.95)}$ & $4.74_{(5.65)}$ & $2.15_{(1.02)}$ & $0.56_{(0.35)}$ & $3.21_{(0.85)}$ & $797_{(572)}$ \\
& resample & $6.11_{(4.76)}$ & $7.41_{(6.76)}$  & $6.74_{(7.48)}$ & $2.25_{(1.04)}$ & $0.58_{(0.33)}$ & $3.16_{(0.88)}$ & $2539_{(662)}$ \\
\midrule
\makecell[l]{Olmo-3.1-32B \\ Think (16k)} & resample & $6.71_{(5.22)}$ & $8.14_{(7.45)}$ & $5.99_{(7.84)}$ & $2.80_{(1.14)}$ & $0.53_{(0.34)}$ & $2.47_{(0.88)}$ & $30939_{(8503)}$ \\
 & creative & $4.79_{(4.07)}$ & $5.25_{(4.97)}$ & $2.08_{(3.17)}$ & $2.83_{(1.22)}$ & $0.46_{(0.37)}$ & $2.45_{(0.80)}$ & $10747_{(3506)}$ \\
 & verbalized & $4.53_{(3.53)}$ & $4.77_{(4.04)}$  & $1.46_{(2.05)}$ & $2.90_{(1.18)}$ & $0.46_{(0.39)}$ & $2.31_{(0.82)}$ & $10002_{(3344)}$ \\
 & iterate & $6.08_{(4.72)}$ & $7.11_{(6.44)}$  & $3.84_{(5.21)}$ & $2.86_{(1.19)}$ & $0.49_{(0.35)}$ & $2.53_{(0.79)}$ & $8435_{(3326)}$ \\
& original & $4.78_{(3.96)}$ & $5.25_{(4.95)}$& $2.16_{(3.22)}$ & $2.81_{(1.25)}$ & $0.45_{(0.38)}$ & $2.40_{(0.82)}$ & $10370_{(3594)}$ \\

\midrule
\makecell[l]{Qwen3-30B\\Instruct} & original & $5.20_{(4.60)}$ & $6.27_{(6.42)}$ & $5.12_{(6.99)}$ & $2.43_{(0.99)}$ & $0.53_{(0.33)}$ & $3.32_{(0.99)}$ & $1938_{(646)}$ \\
 & creative & $5.34_{(4.56)}$ & $6.45_{(6.47)}$ & $3.87_{(6.26)}$ & $2.35_{(0.92)}$ & $0.52_{(0.32)}$ & $3.47_{(1.05)}$ & $2563_{(1074)}$ \\
 & verbalized & $3.66_{(3.46)}$ & $3.96_{(4.08)}$ & $1.77_{(2.70)}$ & $2.43_{(1.08)}$ & $0.47_{(0.38)}$ & $3.41_{(0.85)}$ & $1508_{(1172)}$ \\
 & iterate & $6.24_{(5.16)}$ & $8.12_{(7.99)}$  & $9.84_{(13.31)}$ & $2.47_{(0.99)}$ & $0.52_{(0.32)}$ & $3.31_{(0.96)}$ & $1933_{(672)}$ \\
& resample & $7.10_{(5.24)}$ & $9.34_{(8.34)}$ & $11.00_{(13.71)}$ & $2.43_{(0.93)}$ & $0.56_{(0.31)}$ & $3.40_{(0.94)}$ & $7834_{(2209)}$ \\

\midrule
Qwen3-32B (16k) & original & $4.69_{(3.88)}$ & $5.08_{(4.64)}$ & $2.11_{(2.38)}$ & $2.77_{(1.06)}$ & $0.53_{(0.40)}$ & $2.84_{(0.69)}$ & $3207_{(1292)}$ \\
 & creative & $4.73_{(3.92)}$ & $5.14_{(4.66)}$   & $2.14_{(2.38)}$ & $2.75_{(1.03)}$ & $0.54_{(0.41)}$ & $2.81_{(0.68)}$ & $3257_{(1409)}$ \\
 & verbalized & $4.03_{(3.53)}$ & $4.26_{(4.05)}$  & $1.53_{(1.85)}$ & $2.85_{(1.13)}$ & $0.51_{(0.43)}$ & $2.64_{(0.73)}$ & $3041_{(1261)}$ \\
 & iterate & $6.38_{(4.71)}$ & $7.52_{(6.48)}$  & $4.10_{(4.34)}$ & $2.79_{(0.98)}$ & $0.58_{(0.36)}$ & $2.91_{(0.66)}$ & $2166_{(1221)}$ \\
& resample & $2.99_{(2.96)}$ & $3.10_{(3.20)}$ & $0.11_{(0.66)}$ & $2.40_{(1.03)}$ & $0.47_{(0.41)}$ & $2.95_{(0.90)}$ & $9646_{(3168)}$ \\

\bottomrule
\end{tabular}
\end{table*}

\begin{table*}
\small
\centering
\rowcolors{2}{white}{lightgray}
\renewcommand{\tabcolsep}{0.55mm}
\caption{Performance on CREATE if we do not include factuality in the quality objective. There is a trade off between factuality and creative utiilty.}
\label{tab:table1_res_no_fact}
\begin{tabular}{llllllll}
\toprule
 & $s_{0.9} $ & $|U|$ & $\sigma$ & $d$ & path length & factuality & num tokens \\
\midrule
Claude-Opus-4.7 (high) & $13.27_{(7.35)}$ & $9.02_{(4.61)}$ & $3.11_{(1.06)}$ & $0.65_{(0.26)}$ & $2.90_{(0.62)}$ & $0.77_{(0.19)}$ & $2074_{(1710)}$ \\
Claude-Opus-4.7 (xhigh) & $15.12_{(8.07)}$ & $11.39_{(5.17)}$ & $3.15_{(1.04)}$ & $0.65_{(0.24)}$ & $2.95_{(0.59)}$ & $0.77_{(0.18)}$ & $4163_{(3534)}$ \\

Claude-Sonnet-4.6 (high) & $19.30_{(9.15)}$ & $20.43_{(9.82)}$ & $2.88_{(0.81)}$ & $0.69_{(0.22)}$ & $3.61_{(0.83)}$ & $0.81_{(0.15)}$ & $12063_{(12955)}$ \\
GPT-5 (med) & $15.40_{(9.00)}$ & $28.52_{(15.12)}$ & $2.61_{(1.04)}$ & $0.56_{(0.24)}$ & $4.08_{(1.16)}$ & $0.86_{(0.13)}$ & $18245_{(5822)}$ \\
GPT-5-mini (high) & $14.54_{(10.29)}$ & $24.07_{(10.91)}$ & $2.50_{(1.05)}$ & $0.54_{(0.28)}$ & $3.60_{(0.90)}$ & $0.86_{(0.19)}$ & $23370_{(5683)}$ \\
GPT-5.4 (high) & $16.26_{(9.22)}$ & $37.28_{(13.11)}$ & $2.67_{(1.04)}$ & $0.56_{(0.23)}$ & $4.19_{(1.24)}$ & $0.84_{(0.14)}$ & $14721_{(4166)}$ \\
GPT-5.5 (high) & $18.02_{(9.84)}$ & $40.17_{(16.05)}$ & $2.70_{(1.01)}$ & $0.60_{(0.23)}$ & $4.15_{(1.25)}$ & $0.83_{(0.14)}$ & $13557_{(4948)}$ \\
Gemini-3-pro & $18.70_{(8.63)}$ & $14.10_{(5.82)}$ & $3.02_{(0.87)}$ & $0.73_{(0.22)}$ & $4.56_{(2.24)}$ & $0.76_{(0.15)}$ & $1764_{(1069)}$ \\
Gemini-3.1-Pro & $13.89_{(7.11)}$ & $8.65_{(3.31)}$ & $2.93_{(1.00)}$ & $0.73_{(0.24)}$ & $3.59_{(1.25)}$ & $0.80_{(0.16)}$ & $3200_{(1669)}$ \\
Olmo-3.1-32B-Think (32k) & $9.55_{(6.45)}$ & $4.90_{(5.18)}$ & $2.99_{(1.16)}$ & $0.51_{(0.33)}$ & $2.48_{(0.91)}$ & $0.66_{(0.30)}$ & $10334_{(3634)}$ \\
Qwen\_Qwen3-32B (32k) & $13.15_{(5.82)}$ & $6.89_{(3.29)}$ & $2.88_{(0.84)}$ & $0.73_{(0.27)}$ & $3.00_{(0.73)}$ & $0.60_{(0.22)}$ & $3225_{(1562)}$ \\
gpt-oss-120b (high) & $16.92_{(6.60)}$ & $12.11_{(3.47)}$ & $2.78_{(0.69)}$ & $0.75_{(0.21)}$ & $3.11_{(0.54)}$ & $0.73_{(0.19)}$ & $4484_{(1321)}$ \\
\bottomrule
\end{tabular}
\end{table*}

\subsection{Significance Testing}\label{sec:stats_testing}
For the 6 models in Figure~\ref{fig:results_1}(a), we test all $\binom{6}{2}=15$ pairwise differences in mean creative utility at patience=1.0. For each pair, we compute per-query utility differences (intersecting on the query indices both models scored) and run a paired bootstrap with 10{,}000 resamples to obtain a 95\% percentile CI on the mean difference and a two-sided p-value. We apply Bonferroni correction across the 15 pairs. 13 of 15 pairs are significant at $\alpha=0.05$. The two exceptions are Claude-Sonnet-4.6 vs.\ GPT-5.5 and Claude-Opus-4.7 (xhigh) vs.\ Gemini-3-pro.


\begin{wrapfigure}{r}{0.45\textwidth}
    \centering
    \includegraphics[scale=0.25,trim=0 25cm 0 0cm]{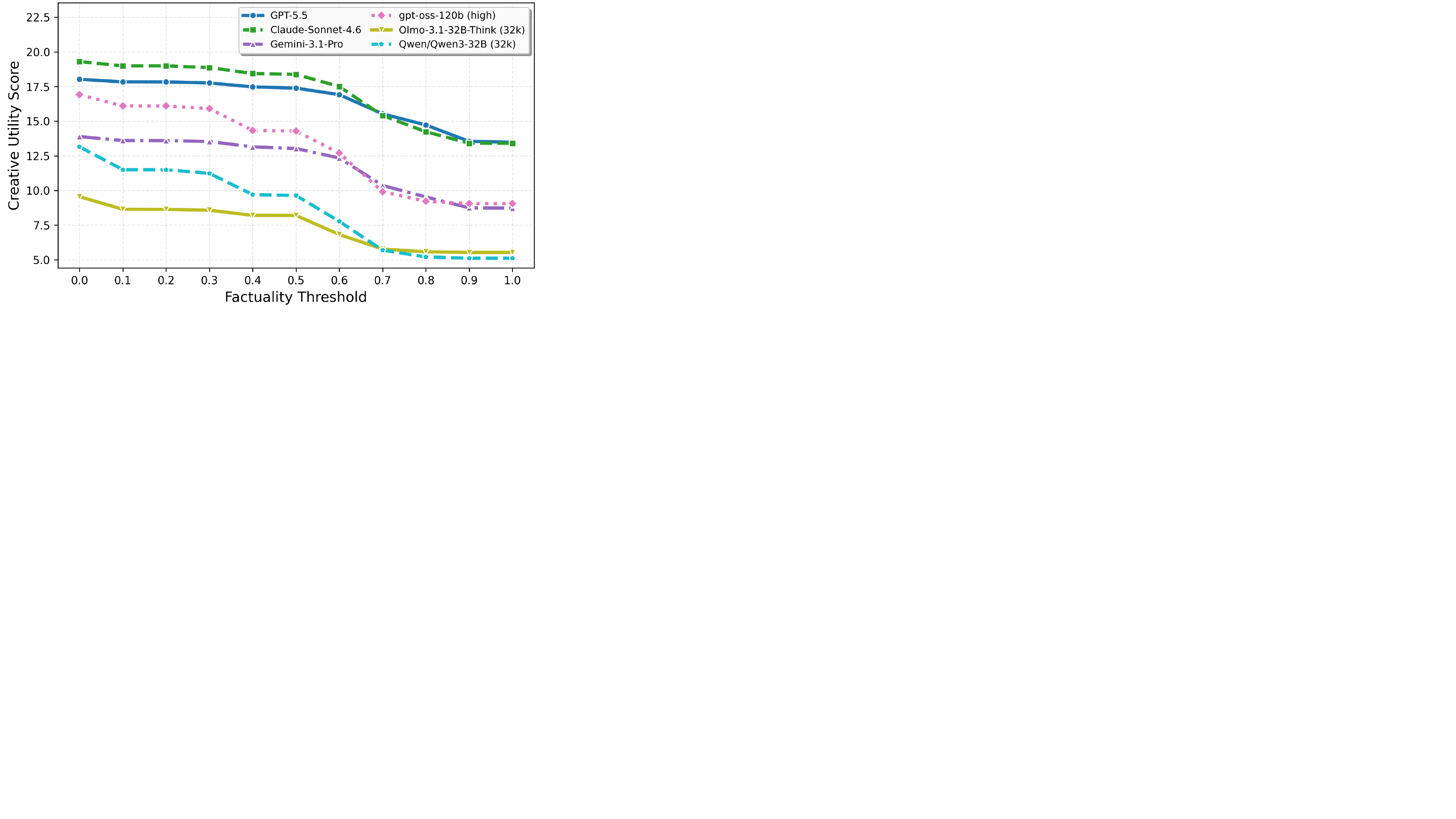}
    \caption{The graph shows how the creative utility (patience=0.9) of a model changes when we include factuality in the objective. Models trade off factuality for utility. }
    \label{fig:tradeoff}
\end{wrapfigure}

\subsection{Quality and Distinctiveness} \label{app:distinctiveness}

Following our initial analysis in Section~\ref{sec:case_study}, we perform an automated experiment to discern the distinctiveness of the generations. Using $\nu$, we analyze the frontier of high-quality responses produced by each model.

We subsample 300 queries from each model and, for each setting, filter out invalid paths, factual inconsistent paths, and paths with quality $\le 3$. For the remaining paths for each query, we calculate each paths' distance from a population of responses for the same query. Avg $\nu$ reflects the average distance to the closest path in the population, and max $\nu$ reflects the best produced instance for each query. 

Table ~\ref{tab:high_distinct} shows that distinctiveness values are similar across models. We particularly focus on prompt variations, and see that on average they do not produce very distinctive responses compared to the base prompt. Interestingly, ``iterate'' has the strongest positive impact here. This metric captures something different than utility: we see that resampling is less useful here, but knowing what a model has generated before and explicitly instructing it to generate different things seems to find more distinct items.

\begin{table}
\small
\centering
\renewcommand{\tabcolsep}{0.7mm}
\caption{We compare how much a model generated path differs from the population of paths produced by all other models. We report the average of the distance between a path and the closest path in the population along with the average strength and then number of paths considered.}
\label{tab:high_distinct}
\begin{tabular}{llrrr}
\toprule
\multicolumn{1}{c}{Model} & Variation & \multicolumn{1}{c}{avg distance} & \multicolumn{1}{c}{avg $\sigma$} & \multicolumn{1}{c}{num} \\
\midrule
Gemini-3-pro & original & 0.04 & 3.77 & 223 \\
\midrule
GPT-5 & original & 0.02 &  3.73 & 224 \\
\midrule
GPT-4.1 & original & 0.03 &  3.76 & 215 \\
 & creative & 0.028 & 3.793 & 216 \\
 & verbalized & 0.020  & 3.820 & 184 \\
 & iterate & 0.053 & 3.600 & 205 \\
 & resample & 0.026 & 3.754 & 219 \\
 \midrule
GPT-5-mini & original & 0.02  & 3.76 & 208 \\
 & creative & 0.03  & 3.75 & 216 \\
 & verbalized & 0.02  & 3.76 & 214 \\
 & iterate & 0.06 & 3.60 & 225 \\
 \bottomrule
\end{tabular}
\vspace{-1.5em}
\end{table}


\subsection{Quality vs Factuality} \label{sec:qualvsfact}

Table \ref{tab:table1_res_no_fact}, we report creative utility without any factuality component of quality, along with the average number of factually correct triples generated by a model. We observe a tradeoff between utility and factuality: GPT-5 achieves lower utility compared to Claude-Sonnet-4.6 but has higher factuality. 


To quantify the utility-factuality tradeoff, we define a factuality filter. We first define $t$ as the fraction of triples in a path that are factually correct. We then vary $t$ and modify Equation~\ref{eq:utility} to compute a factuality-adjusted utility:
\vspace{-1em}
\begin{equation*}
\small
\begin{aligned}
s(U, t) = \max_{\tau} \sum_{i=1}^{|U|}
& \gamma^{i-1}
\, \mathbb{I}\!\left[q\!\left(u_{\tau(i)}\right) > t \right]
& \times \sigma\!\left(u_{\tau(i)}\right)
\min_{j<i} d\!\left(u_{\tau(i)}, u_{\tau(j)}\right),
\end{aligned}
\end{equation*}
\vspace{-1em}

Note, instead of having a unified quality metric as defined in Equation~\ref{eq:quality}, we consider factuality to be another dimension for this analysis and use $\sigma$ as the quality measure.

Figure \ref{fig:tradeoff} shows a consistent drop in creative utility the factuality cutoff increases, as expected. At the most lenient, when we consider all valid paths generate by the models, we observe Sonnet-4.6 achieving the highest creative utility, followed by GPT-5.5, then gpt-oss-120b close behind. But, at the strictest setting, only retaining paths that are entirely factual, we observe GPT-5.5 and Sonnet-4.6 performing the best, while open-source models have a substantial drop, revealing a large gap. Note that because many paths have 2 or 3 relations, the steepest drop-off is from a threshold of 0.5 to 0.7.

\subsection{Search Trace} \label{sec:search_trace}

We use Prompt \ref{prompt:search_trace} to extract entities and relations from a search trace using gpt-4.1-mini-2025-04-14. Table \ref{tab:think} summarizes the number of entities, relations and triples explored by the model in its reasoning trace.

\begin{figure}
    \centering
    \includegraphics[scale=0.45,trim=0 0 0 0cm]{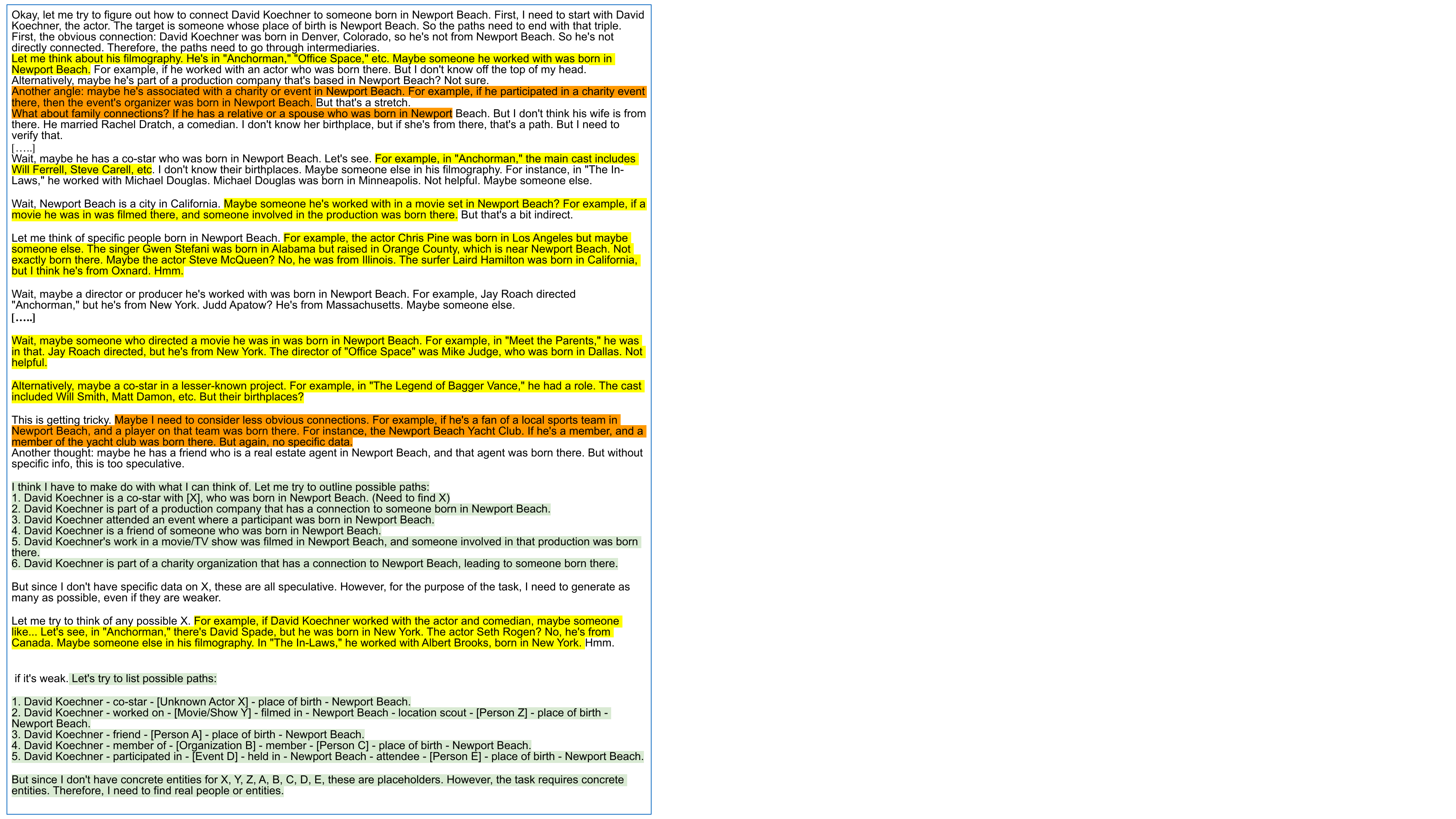}
    \caption{Example of a reasoning chain for a query `\textit{What are possible connections between David Koechner and someone who was born in Newport Beach}?'}
    \label{fig:trace_example}
\end{figure}

Figure \ref{fig:trace_example} shows an example of a reasoning trace. We see various strategies the model uses to brainstorm connections, like enumeration, cross domain connections and template based search. We also see a lot of repetitions of entities and search paths, indicating not optimal use of tokens for this task and thus identifying a gap.

\begin{wraptable}{r}{0.5\textwidth}
\small
\centering
\renewcommand{\tabcolsep}{0.7mm}
\caption{Number of entities (e) and relations (r) explored by thinking models in their reasoning traces.}
\label{tab:think}
\begin{tabular}{lrrr}
\toprule
\multicolumn{1}{c}{Model} & \multicolumn{1}{c}{ r} & \multicolumn{1}{c}{ e}  \\
\midrule
Olmo-3.1-32B / 16k & 12.04 & 18.83 \\
Olmo-3.1-32B  / 32k & 12.22 & 20.05 \\
Qwen3-32B / 16k & 15.12 & 27.33  \\
Qwen3-32B / 32k & 14.81 & 26.18  \\ 
gpt-oss-120b / low & 24.23 & 28.97 \\ 
gpt-oss-120b /high & 24.73 & 27.99  \\
\bottomrule
\end{tabular}
\end{wraptable}

\section{Tests for human creativity and diversity}
\label{sec:rat_aut_etc}

Human creativity is often assessed using tasks like Remote Associations Test (RAT) \cite{mednick1968remote}, Alternate Uses Task (AUT) \cite{guilford1978alternate}, and Diversity Association Task (DAT) \cite{olson2021naming}. In this section, we show that these tests are saturated for frontier models and new benchmarks are needed.

We describe the tests along with the evaluation criteria:

\paragraph{Remote Associations Test (RAT)} This test consists of three common stimulus words that appear to be unrelated. The subject must think of a fourth word that is somehow related to each of the first three words. 

\paragraph{Alternate Uses Task (AUT)} This test measures creativity of a subject by asking them to list as many unique uses for a common object (e.g. a brick or a paperclip) as possible. It evaluates four key areas generally, fluency, originality, flexibility and elaboration. In this work we only focus on originality \cite{wenger2025we}.

\paragraph{Diversity Association Task (DAT)} This task involves naming ten nouns that differ as much as possible from each other \cite{wenger2025we}.

For RAT, we calculate the accuracy. For AUT and DAT we follow the originality calculations according to prior work \cite{wenger2025we}. For AUT, originality is how novel the response is compared to the prompt. For DAT, originality is how different a pair of words are in the response. For both of these tests, we use glove embeddings \cite{pennington2014glove} and calculate cosine distance as measure of novelty. Since AUT only has 5 query words, we run each of these 10 times, to get an evaluation set of 50 instances. Similarly for DAT, we sample 50 responses from the model.

Table~\ref{app:aut-qualitative-examples} shows examples of instances for each of the tests along with model responses and the quantitative scores. In Table~\ref{app:alternate_benchmarks} we summarize the performance for GPT-4.1-mini and GPT-5-mini on these three benchmarks. We see that flagship reasoning models achieve very high accuracy on RAT. For the diversity benchmarks AUT and DAT, we see both models scoring high on originality.

\begin{table}[t]
\centering
\caption{Human creativity test, example, number of instances, and LLM performance on these tests. We only look at average accuracy and average originality for RAT and AUT/DAT respectively. }
\label{app:alternate_benchmarks}
\begin{tabular}{ccccc}
\toprule
    Test & Num points & Metric & GPT-4.1-mini & GPT-5-mini \\
\midrule 
RAT &  212 & accuracy & 0.48 (0.5) & 0.86 (0.35)\\ 
\midrule 
AUT &  \makecell{5 words / \\ 10 samples each} & originality & 0.81 (0.06) & 0.80 (0.09) \\
\midrule
DAT & 50 samples & originality & 0.84 (0.01) & 0.82 (0.03) \\

\bottomrule
\end{tabular}
\end{table}

\begin{table}[htbp]
\centering
\caption{Qualitative examples for GPT-5-mini (gpt-5-mini-2025-08-07). AUT and DAT scores are originality; RAT score is accuracy.}
\label{app:aut-qualitative-examples}
\small
\begin{tabularx}{\textwidth}{l|X|X|c}
\toprule
Test & Instruction (full prompt) & Model response & Score \\
\midrule
RAT &
Find a fourth word that is related to all three of the following words: cottage / swiss / cake Respond with only the single word answer, nothing else. &
cheese &
1.0 \\
& & \\ 
AUT &
Imagine that someone gives you a book. Write down as many uses as you can think of for this object, up to 10 uses. Please list the uses as words or phrases (single word answers are ok), separated by semicolons. Do not write anything besides your proposed uses. &
Reading; Doorstop; Paperweight; Coaster; Gift; Kindling; Art canvas; Pressing flowers; Prop for photos; Shelf divider &
0.83 \\
& & \\ 
DAT &
Instructions: Please enter 10 words that are as different from each other as possible, in all meanings and uses of the words. Rules: 1. Only single words in English. 2. Only nouns (e.g., things, objects, concepts). 3. No proper nouns (e.g., no specific people or places). 4. No specialised vocabulary (e.g., no technical terms). 5. Think of the words on your own (e.g., do not just look at objects in your surroundings). 6. Complete this task in less than four minutes. 7. Return just the list of words, separated by commas, and do not include any other content. &
mountain, idea, apple, justice, elephant, ship, song, brick, sand, cloud &
0.88 \\
\bottomrule
\end{tabularx}
\end{table}

\begin{prompt}[float*,
width=\textwidth,
title={\footnotesize\texttt{Prompt \thetcbcounter: Strength Prompt}}, 
label=prompt:strength_prompt,
breakable]
Goal:
\\
Given a knowledge graph path consisting of multiple triples of the form (subject, predicate, object), estimate — FOR EACH TRIPLE — the number of members in the LARGER of two possible classes defined by the predicate.
\\
Formal Definitions (apply to each triple (s, p, o)):
Define TWO classes.
\\
Class A (Subject-variation class): \\
    \{\{ x | (x, p, o) is true \}\}\\
This is the set of all possible SUBJECTS x that could replace s while keeping (p, o) fixed.\\
The given subject s counts as ONE MEMBER of this class.\\
\\
Class B (Object-variation class):\\
    \{\{ y | (s, p, y) is true \}\}\\
This is the set of all possible OBJECTS y that could replace o while keeping (s, p) fixed.\\
The given object o counts as ONE MEMBER of this class.\\
\\
Directional Discipline (CRITICAL):\\
- You MUST evaluate both Class A and Class B independently for each triple.
- NEVER collapse or reinterpret the relation.\\
- For directional predicates (e.g., 'influenced', 'taught', 'founded', 'won', 'received'):\\
    * Class A asks: who/what stands in relation (predicate) TO the object? \\
    * Class B asks: who/what the subject stands in relation (predicate) TO? \\
- Always respect the original direction of the predicate. \\
\\
Procedure (apply to EACH triple in the path independently):\\
1. Identify Class A from (p, o).\\
2. Identify Class B from (s, p).\\
3. Use factual knowledge and reasonable estimation to estimate the size of EACH class.\\
4. Select the LARGER of the two estimated class sizes.\\
5. Explain your reasoning step by step for both classes.\\
\\
Output instructions:\\
- Return valid JSON only.\\
- The output must be a JSON list, with one object per triple, in the same order as the path.\\
- Each object must contain exactly two keys:\\
  - ``explanation": a natural-language explanation for that triple’s Class A and Class B reasoning.\\
  - ``judgment": a single integer representing the estimated size of the LARGER class.\\
- Do not include any text outside the JSON list.\\
- Do not include markdown or extra keys.\\
- Each judgment must be a single integer (no commas, no extra text).\\
The output must exactly match this schema:\\

  \{\{
    ``explanation": ``string",
    ``judgment": integer
  \}\}
\\

==== INPUT ====

Path: \{ \paramnorm{path} \}

\end{prompt}

\begin{prompt}[float*,
width=\textwidth,
title={\footnotesize\texttt{Prompt \thetcbcounter: Factuality Prompt}}, 
label=prompt:factuality_prompt,
breakable]
You are a fact-checking expert evaluating a multi-hop factual path for factual accuracy and logical validity.
\\

The path consists of an ordered list of triples.
Each triple has the form: (subject, relation, object)
\\

Your task is to evaluate EACH triple in the given path and determine whether it is hallucinated or not. Use the path only for entity disambiguation and contextual grounding. Do not infer unsupported facts.
\\

Evaluation criteria (apply independently to each triple):
\\

1. Entity existence and disambiguation:\\
   - Verify that both the subject and object are real, identifiable entities (people, places, organizations, events, or concepts).\\
   - Use the path context only when necessary to disambiguate entities.\\
   - If either entity is fabricated or incorrectly identified, mark the triple as hallucinated.\\
\\

2. Relationship validity:\\
   - Verify that the stated relation describes a real, historically valid, currently valid, or plausibly valid relationship between the subject and object.\\
   - If the relationship is plausible and could reasonably have occurred, even if not directly verifiable, mark it as not hallucinated.\\
   - If it is niche information about the subject and only trivia heads would know compared to common knowledge then do not mark it as hallucinated.\\
\\

3. Hallucination detection:\\
   - Mark the triple as hallucinated if:\\
     - An entity is fabricated or does not exist\\
     - The relation is fabricated, nonsensical, or incorrect\\
     - The asserted connection clearly contradicts well-known facts\\
     - The triple introduces made-up properties or roles\\
\\

4. Directionality handling:
   - If the relation is ambiguous or commonly bidirectional, evaluate both directions.
   - If either direction corresponds to a true or plausible relationship, mark the triple as not hallucinated.
\\

Output instructions:
- Return valid JSON only.\\
- Include exactly two keys: ``explanation" and ``judgments".\\
- Under ``explanation", provide a natural-language explanation evaluating the triples in the path.\\
- Under ``judgments", provide the final hallucination judgment for each triple as a list, in the same order as the path.\\
- Do not include any text outside the JSON object.\\
- Do not include markdown or extra keys.\\
\\

The output must exactly match this schema:

\{\{
  ``explanation": ``string",
  ``judgments": [``hallucinated" | ``not hallucinated", ...]
\}\}
\\
Input:\\ 

Path: \{ \paramnorm{path} \}
\end{prompt}

\begin{prompt}[float*,
width=\textwidth,
title={\footnotesize\texttt{Prompt \thetcbcounter: Base prompt used in CREATE }}, 
label=prompt:base_prompt,
breakable]

Query: \{ \paramnorm{query} \} \\

Task: Identify how two real-world entities are connected by producing MANY connection paths. A connection path is a sequence of factual triples (head, relationship, tail) forming a continuous chain that begins with one entity and ends with a required target condition.\\

You MUST generate as many distinct valid paths as possible. Within each individual path, prefer STRONG connections (highly exclusive, specific relationships). Across the full set of paths, maintain DIVERSITY: include both popular/well-known connections and less well-known “trivia” connections, and avoid over-concentrating on the most obvious domain (e.g., for a celebrity, do not only use their main profession—add distinct non-professional connections when available).\\
\\
Path definition:
- Every path MUST start with the head entity: `[entity a]' \\
- Every path MUST end with a triple whose relationship is '[rel b]' and whose tail entity is '[entity b]' \\
- Paths may be direct or indirect and may include one or more intermediate entities \\

Rules and quality constraints: \\
- Entities must be concrete, real-world entities only (people, organizations, works, places, genes, diseases, species, etc.). No abstract concepts. \\
- Do not ask follow-up questions; respond using the best available factual knowledge. \\
- Temporal connections are allowed (relationships may span different historical periods). \\
- Disambiguation is required: use canonical names and qualifiers where necessary (e.g., 'Michael Jordan (basketball)'). \\
- If multiple canonical entities share the same name, explore ALL of them explicitly where relevant. \\

Deduplication: \\
- Do not repeat the same path. \\
- Do not repeat the same triple within a single path. \\
- Prefer paths that are meaningfully different (different intermediate nodes and/or different relationships), not trivial rephrasings. \\

Coverage and diversity: \\
- Generate as many distinct valid paths as you can. \\
- Explore a broad range of relationship types for '[entity a]'. \\
- Include BOTH: \\
  (a) strong/obvious connections (the first things most people would think of), AND \\
  (b) less well-known but still factual connections (“trivia”) that are distinct from the popular ones. \\
- After you have produced several paths in a dominant domain (e.g., movies/acting for an actor), actively search for other distinct domains (e.g. philanthropy) when possible. \\

Relationship quality guidance: \\
- Prefer strong, specific, and distinctive relationships. \\
  - Strong = highly exclusive (e.g., parent/child, founder-of, spouse, authored, CEO-of, member-of a small group). \\
  - Weaker = shared broad attributes (e.g., “attended”, “lives in”, “worked on” in very large productions). \\
- In each individual path, prioritize strong links early in the chain when possible. \\
- Across paths, start with strong + distinctive paths, then include progressively more general/weaker but still valid paths to maximize coverage. \\
\\
Output requirements (strict): \\
- Return ONLY a JSON object wrapped in answer tags. Do not include any explanatory text. \\
- The JSON object must use integer keys starting from 1. \\
- Each triple must be of the form: (head entity, relationship, tail entity). \\
- Relationship strings must be 1–3 words. \\
- If no valid path exists, return an empty JSON object. \\
\\
Enumerate all distinct valid connection paths that satisfy the above constraints. \\

\end{prompt}

\begin{prompt}[float*,
width=\textwidth,
title={\footnotesize\texttt{Prompt \thetcbcounter: Verbalized Sampling Prompt}}, 
label=prompt:verbalized_sampling,
breakable]

Query: \{ \paramnorm{query} \}
\\

Task: Identify how two real-world entities are connected by producing MANY connection paths. A connection path is a sequence of factual triples (head, relationship, tail) forming a continuous chain that begins with one entity and ends with a required target condition.
\\

You MUST generate as many distinct valid paths as possible. Within each individual path, prefer STRONG connections (highly exclusive, specific relationships). Across the full set of paths, maintain DIVERSITY: include both popular/well-known connections and less well-known “trivia” connections, and avoid over-concentrating on the most obvious domain (e.g., for a celebrity, do not only use their main profession—add distinct non-professional connections when available).
\\

Path definition:
- Every path MUST start with the head entity: 'Morton Gould'
- Every path MUST end with a triple whose relationship is 'discography' and whose tail entity is 'Benny Carter discography'
- Paths may be direct or indirect and may include one or more intermediate entities
\\

Rules and quality constraints:
- Entities must be concrete, real-world entities only (people, organizations, works, places, genes, diseases, species, etc.). No abstract concepts.
- Do not ask follow-up questions; respond using the best available factual knowledge.
- Temporal connections are allowed (relationships may span different historical periods).
- Disambiguation is required: use canonical names and qualifiers where necessary (e.g., 'Michael Jordan (basketball)').
- If multiple canonical entities share the same name, explore ALL of them explicitly where relevant.
\\

Deduplication:
- Do not repeat the same path.
- Do not repeat the same triple within a single path.
- Prefer paths that are meaningfully different (different intermediate nodes and/or different relationships), not trivial rephrasings.
\\

Coverage and diversity:
- Generate as many distinct valid paths as you can.
- Explore a broad range of relationship types for 'Morton Gould'.
- Include BOTH:
(a) strong/obvious connections (the first things most people would think of), AND
(b) less well-known but still factual connections (“trivia”) that are distinct from the popular ones.
- After you have produced several paths in a dominant domain (e.g., movies/acting for an actor), actively search for other distinct domains (e.g. philanthropy) when possible.
\\

Relationship quality guidance:
- Prefer strong, specific, and distinctive relationships.
- Strong = highly exclusive (e.g., parent/child, founder-of, spouse, authored, CEO-of, member-of a small group).
- Weaker = shared broad attributes (e.g., “attended”, “lives in”, “worked on” in very large productions).
- In each individual path, prioritize strong links early in the chain when possible.
- Across paths, start with strong + distinctive paths, then include progressively more general/weaker but still valid paths to maximize coverage.
\\

- For each path, assign a normalized confidence score in [0.0,1.0] representing the relative likelihood that a knowledgeable person would recognize or know this connection.
Higher scores should correspond to more direct, typical, or well-known relationships, while lower scores should correspond to more indirect, obscure, or atypical relationships.
The confidence scores across all generated paths must sum to exactly 1.0, and the paths should be ordered from highest to lowest confidence.
\\

Output requirements (strict):
- Return ONLY a JSON object wrapped in <answer> tags. Do not include any explanatory text.
\\

- The JSON object must use integer keys starting from 1.
\\

- Each integer key maps to an object with:
``path\_probability'': a float in the range $[0.0,1.0]$, rounded to two decimal places, representing the normalized likelihood of the path relative to the other paths, such that the probabilities across all paths sum to 1.0.
\\
``path'': a list of triples of the form (head entity, relationship, tail entity).
- Each triple must be of the form: (head entity, relationship, tail entity).
\\
- Relationship strings must be 1–3 words.
\\
- If no valid path exists, return an empty JSON object.
\\

Enumerate all distinct valid connection paths that satisfy the above constraints.

\end{prompt}

\begin{prompt}[float*,
width=\textwidth,
title={\footnotesize\texttt{Prompt \thetcbcounter: One round of Iterative prompting}}, 
label=prompt:iterative,
breakable]
\{ ``content'': base prompt, ``role'': ``user''\},

\{
``content": [answer from previous iteration], ``role": ``assistant" \},

\{
``content": ``I am restating the query: \{ \paramnorm{query} \}

Give me more/different associations than the answers you gave in the previous response. If your previous response is empty, then try again.

Output requirements (strict): \\
- Return ONLY a JSON object wrapped in answer tags. Do not include any explanatory text. \\
- The JSON object must use integer keys starting from 1. \\
- Each triple must be of the form: (head entity, relationship, tail entity). \\
- Relationship strings must be 1–3 words. \\
- If no valid path exists, return an empty JSON object. \\
\\
Enumerate all distinct valid connection paths that satisfy the above constraints.", ``role": ``user" \}

\end{prompt}

\begin{prompt}[float*,
width=\textwidth,
title={\footnotesize\texttt{Prompt \thetcbcounter: Prompt for extracting entities and relations from a reasoning trace}}, 
label=prompt:search_trace,
breakable]

You are given a search reasoning trace: a long, free-form explanation of how an answer was derived, including candidate facts, entities, and relationships.

Your task is to extract a structured JSON object with two keys: ``relation" and ``entities".

========================
\#\#\# 1. Output Format
========================

Return **only** a single JSON object with this exact top-level structure:

\{\{
  ``relation": (``relation1",``relation2",...),
  ``entities": (``entity1",``entity2",...)
\}\}

No extra keys, no comments, no explanations, no text outside the JSON, and no trailing commas.

========================
\#\#\# 2. Key Definitions
========================

\#\#\#\# ``relation" \\
- An array of **strings**.\\
- Each string is a **relation / predicate phrase** that appears explicitly or is clearly implied in the trace, describing how two entities can be linked.\\
- Include:\\
  - Verb/verb-phrase relations (e.g., ``worked at", ``served in", ``attended (alma mater)"). \\
  - Noun-phrase relations that behave as connectors (e.g., ``commander-in-chief"). \\
- **Do not** include: \\
  - Full sentences.\\
  - Dates, numbers, or descriptions unrelated to relations.\\
  - Entity names.\\
\\

\#\#\#\# ``entities" \\
- An array of **strings**. \\
- Each string is an **entity or concept** mentioned or implied in the reasoning trace. \\
- Include: \\
  - People (e.g., ``Joe Biden", ``John Adams", ``Abigail Adams", ``George Washington"). \\
  - Institutions and places (e.g., ``White House", ``U.S. Senate", ``National Archives"). \\
  - Roles or collective bodies (e.g., ``U.S. Presidency", ``First Lady of the United States"). \\
  - Abstract but named items (e.g., ``List of U.S. Presidents", ``Media portrayals (e.g., John Adams miniseries)").\\
- **Do not** include:\\
  - Relations (those go into ``relation").\\
  - Dates, numbers, or purely descriptive adjectives.\\
  - Full sentences.\\
\\

========================
\#\#\# 3. General Instructions
========================
\\

1. **Scan the entire reasoning trace** and extract:\\
   - All relation phrases → ``relation" \\
   - All entity names or conceptual entities → ``entities"\\
2. **Do not deduplicate** items within each list. Generate all entities and relations that appear in the trace as they appear in the trace. Do not remove duplicates.\\
3. Normalize entity and relation names to lowercase, remove extra spaces, and remove punctuation.\\
3. **Do not infer facts** that are not mentioned or clearly implied.\\
4. **Return only valid JSON**, with:\\
   - Double quotes around strings\\
   - Arrays for both keys \\
   - No other output\\
\\

========================
\#\#\# 4. Input to process
========================
\\

\{ \paramnorm{trace} \}
\end{prompt}

\clearpage
\newpage
\section*{NeurIPS Paper Checklist}

\begin{enumerate}

\item {\bf Claims}
    \item[] Question: Do the main claims made in the abstract and introduction accurately reflect the paper's contributions and scope?
    \item[] Answer: \answerYes{}  
    \item[] Justification: The main claims are in the abstract and Section 1.

\item {\bf Limitations}
    \item[] Question: Does the paper discuss the limitations of the work performed by the authors?
    \item[] Answer: \answerYes{}  
    \item[] Justification: Limitations are discussed in the Appendix.

\item {\bf Theory assumptions and proofs}
    \item[] Question: For each theoretical result, does the paper provide the full set of assumptions and a complete (and correct) proof?
    \item[] Answer: \answerNA{} 
    \item[] Justification: Our work does not contain theoretical results

    \item {\bf Experimental result reproducibility}
    \item[] Question: Does the paper fully disclose all the information needed to reproduce the main experimental results of the paper to the extent that it affects the main claims and/or conclusions of the paper (regardless of whether the code and data are provided or not)?
    \item[] Answer: \answerYes{} 
    \item[] Justification: We will release the benchmark and the evaluation code. All evaluation prompts are given in Appendix \ref{sec:eval_prompts}. All inference prompts are given in \ref{app:inference_prompts}. The inference details are given in the experimental setup in Section \ref{sec:experimental_setup} and Appendix \ref{app:model_details}.

\item {\bf Open access to data and code}
    \item[] Question: Does the paper provide open access to the data and code, with sufficient instructions to faithfully reproduce the main experimental results, as described in supplemental material?
    \item[] Answer: \answerYes{} 
    \item[] Justification: We will release the benchmark and the evaluation code with proper documentation.

\item {\bf Experimental setting/details}
    \item[] Question: Does the paper specify all the training and test details (e.g., data splits, hyperparameters, how they were chosen, type of optimizer) necessary to understand the results?
    \item[] Answer: \answerYes{} 
    \item[] Justification: All evaluation prompts are given in Appendix \ref{sec:eval_prompts}. All inference prompts are given in \ref{app:inference_prompts}. The inference details are given in the experimental setup in Section \ref{sec:experimental_setup} and Appendix \ref{app:model_details}. We also discuss the data construction process in Section \ref{sec:benchmark} and Appendix \ref{app:benchmark} and the human evaluation process in Section \ref{sec:case_study} and Appendix \ref{sec:human_eval}.

\item {\bf Experiment statistical significance}
    \item[] Question: Does the paper report error bars suitably and correctly defined or other appropriate information about the statistical significance of the experiments?
    \item[] Answer: \answerYes{} 
    \item[] Justification: We use paired bootstrap tests to check if the difference in performance between the models is significant or not. We report the results in Appendix \ref{sec:stats_testing}. We also mark the confidence intervals on the results for prompt variations in Figure \ref{fig:pv_variations}.

\item {\bf Experiments compute resources}
    \item[] Question: For each experiment, does the paper provide sufficient information on the computer resources (type of compute workers, memory, time of execution) needed to reproduce the experiments?
    \item[] Answer: \answerYes{} 
    \item[] Justification: We include details about the compute resources in Appendix \ref{app:model_details}.
    
\item {\bf Code of ethics}
    \item[] Question: Does the research conducted in the paper conform, in every respect, with the NeurIPS Code of Ethics \url{https://neurips.cc/public/EthicsGuidelines}?
    \item[] Answer: \answerYes{} 
    \item[] Justification: The research conducted in the paper conform in every respect with the NeurIPS Code of Ethics.

\item {\bf Broader impacts}
    \item[] Question: Does the paper discuss both potential positive societal impacts and negative societal impacts of the work performed?
    \item[] Answer: \answerNA{} 
    \item[] Justification: This work presents a dataset for benchmarking LLMs in their ability to do creative ideation. As a benchmark, its purpose is primarily to understand these capabilities, and we do not foresee any major risks.

    
\item {\bf Safeguards}
    \item[] Question: Does the paper describe safeguards that have been put in place for responsible release of data or models that have a high risk for misuse (e.g., pre-trained language models, image generators, or scraped datasets)?
    \item[] Answer: \answerNA{} 
    \item[] Justification: We do not believe our data has a high risk of misuse, and we do not release any models.

\item {\bf Licenses for existing assets}
    \item[] Question: Are the creators or original owners of assets (e.g., code, data, models), used in the paper, properly credited and are the license and terms of use explicitly mentioned and properly respected?
    \item[] Answer: \answerYes{} 
    \item[] Justification: We mention the license in Appendix \ref{app:benchmark}. 

\item {\bf New assets}
    \item[] Question: Are new assets introduced in the paper well documented and is the documentation provided alongside the assets?
    \item[] Answer: \answerYes{} 
    \item[] Justification: We will upload the benchmark and code with documentation.

\item {\bf Crowdsourcing and research with human subjects}
    \item[] Question: For crowdsourcing experiments and research with human subjects, does the paper include the full text of instructions given to participants and screenshots, if applicable, as well as details about compensation (if any)? 
    \item[] Answer: \answerNA{} 
    \item[] Justification: This work does not involve crowdsourcing nor research with human subjects.

\item {\bf Institutional review board (IRB) approvals or equivalent for research with human subjects}
    \item[] Question: Does the paper describe potential risks incurred by study participants, whether such risks were disclosed to the subjects, and whether Institutional Review Board (IRB) approvals (or an equivalent approval/review based on the requirements of your country or institution) were obtained?
    \item[] Answer: \answerNA{} 
    \item[] Justification: The paper does not involve any human subjects research.

\item {\bf Declaration of LLM usage}
    \item[] Question: Does the paper describe the usage of LLMs if it is an important, original, or non-standard component of the core methods in this research? Note that if the LLM is used only for writing, editing, or formatting purposes and does \emph{not} impact the core methodology, scientific rigor, or originality of the research, declaration is not required.
    \item[] Answer: \answerNA{} 
    \item[] Justification: The core method developed in this research does not involve LLMs in any important, original or non-standard components.

\end{enumerate}

\end{document}